\documentclass[10pt, a4paper]{article}

\usepackage{lrec-coling2024}

\usepackage{amsmath, booktabs}
\usepackage{bbding}
\usepackage{graphicx}
\usepackage{amssymb, pifont}
\usepackage{subcaption}
\usepackage{array}
\usepackage{comment}

\newcommand{\xmark}{\ding{55}}%
\newcommand{\cmark}{\ding{51}}%

\newcommand{\enna}{$^{\dagger}$}
\newcommand{\equalcontrib}{$^{*}$}
\newcommand{\poli}{$^{\ddagger}$}
\newcommand{\unipa}{$^{\dagger\dagger}$}

\newcommand\blfootnote[1]{%
  \begingroup
  \renewcommand\thefootnote{}\footnote{#1}%
  \addtocounter{footnote}{-1}%
  \endgroup
}

\title{Speech Analysis of Language Varieties in Italy}

\name{
   Moreno La Quatra\equalcontrib \enna, 
   Alkis Koudounas\equalcontrib \poli,
   Elena Baralis\poli, 
   Sabato Marco Siniscalchi\unipa
}

\address{
   \enna Kore University of Enna, 
   \poli Politecnico di Torino,
   \unipa Università degli Studi di Palermo \\
   \enna moreno.laquatra@unikore.it,  
   \poli \{name.surname\}@polito.it, 
   \unipa sabatomarco.siniscalchi@unipa.it \\
   }

\abstract{
Italy exhibits rich linguistic diversity across its territory due to the distinct regional languages spoken in different areas. 
Recent advances in self-supervised learning provide new opportunities to analyze Italy's linguistic varieties using speech data alone. This includes the potential to leverage representations learned from large amounts of data to better examine nuances between closely related linguistic varieties. 
In this study, we focus on automatically identifying the geographic region of origin of speech samples drawn from Italy's diverse language varieties. 
We leverage self-supervised learning models to tackle this task and analyze differences and similarities between Italy’s regional languages.
In doing so, we also seek to uncover new insights into the relationships among these diverse yet closely related varieties, which may help linguists understand their interconnected evolution and regional development over time and space.
To improve the discriminative ability of learned representations, we evaluate several supervised contrastive learning objectives, both as pre-training steps and additional fine-tuning objectives. Experimental evidence shows that pre-trained self-supervised models can effectively identify regions from speech recording. Additionally, incorporating contrastive objectives during fine-tuning improves classification accuracy and yields embeddings that distinctly separate regional varieties, demonstrating the value of combining self-supervised pre-training and contrastive learning for this task.
 \\ \newline  }

\begin{document}

\maketitleabstract

\section{Introduction}
\label{sec:intro}

\blfootnote{\equalcontrib{} Both authors contributed equally to this work.}

% \begin{itemize}
%     \item Provide an overview of the linguistic diversity in Italy.
%     \item Highlight the significance of studying regional dialects.
%     \item Mention potential applications in linguistics and culture preservation.
%     \item Mention the proposed methodology and its novelty.
%     \item What are the expected outcome of our analysis?
% \end{itemize}

The study of linguistic variation in the form of varieties, dialects, and regional languages is becoming an increasingly important area of research within the field of natural language processing~\cite{zampieri2020natural}. 
Analysis of the differences between related language forms at various levels, from lexical features to grammatical structures to phonological patterns, provides unique opportunities to advance NLP techniques. 
Examining the nuances between closely related varieties can help improve systems' abilities to handle diverse linguistic inputs and gain a more comprehensive understanding of a language landscape. 

Italy represents a fascinating case study for investigating linguistic variety within a single nation~\cite{maiden2006dialects}. 
This country exhibits a rich diversity of local languages concentrated within its geographic borders~\cite{ramponi2022nlp}. 
This extensive linguistic heterogeneity arises from Italy's unique historical and cultural influences over time.
Furthermore, the integration and use of regional languages alongside Standard Italian have contributed additional layers of complexity to the nation's linguistic landscape. 
% However, within this mosaic of linguistic differences also lies an opportunity to better understand the cultural and social forces that shape language change and variation.
Analyzing Italy's diverse yet interrelated regional languages allows for exploring the cultural and social factors shaping the development of linguistic variations across communities over time.

This paper explores linguistic variation within Italy using data-driven acoustic analysis of speech signals without resorting to intermediate textual transcriptions.
We analyze the feasibility of automatically determining the geographic origin of speech samples %drawn from Italy’s regional language varieties 
based solely on their acoustic properties.
% Specifically, we design a data-driven approach applying modern machine learning techniques directly to speech signals to classify regional varieties and assess how closely related language varieties % well-related forms 
% can be discerned from the acoustic modality alone.

In this work, we refer to this task as region or language variety identification rather than dialect classification. 
This is because the regional languages, such as those spoken across different parts of Italy, are not necessarily distinct dialects of Standard Italian~\citep[\textit{inter alia}]{avolio2009lingue, ramponi2022nlp}. 
Instead, they represent linguistic variations developed locally within different geographical areas and speech communities in Italy. The term \textit{``language variety''} more accurately describes these regional forms of the language. By using this terminology, we aim to avoid potential limitations or ambiguities of the term \textit{``dialect''}, while still focusing on the core task. % of identifying the origin of a speech sample based on linguistic characteristics. 
The regional varieties are not dialects of Italian \textit{per se}~\cite{berruto2005dialect}, but represent locally developed forms of language usage.

%To analyze the diverse languages spoken across Italy, we leverage the VIVALDI dataset~\cite{vivaldi}, an extensive collection of speech recordings from cities throughout the country. 
To address the region identification task, we leverage VIVALDI~\cite{vivaldi}, an extensive collection of speech recordings from cities throughout the country. 
Uniquely, this dataset contains samples of local language varieties spoken in their native form across Italian regions.
%We explore using self-supervised models to analyze this dataset and gain new insights into Italy's rich linguistic landscape.

% We also investigate the use of contrastive learning objectives to both enhance model performance and representation quality.
From a methodological perspective, we investigate the use of contrastive learning objectives to both enhance the model's ability to accurately identify the geographic region of speech samples, as well as improve the quality of the learned acoustic representations.
While contrastive learning is commonly used in a self-supervised setting~\cite{chen2020simple, al2021clar, giorgi2021declutr}, recent works have shown it can also be effective when applied in a supervised manner~\cite{khosla2020supervised, fu22_interspeech}.
Specifically, we apply contrastive losses: (i) as an additional pre-training step, (ii) as an auxiliary loss during fine-tuning, and (iii) by combining the two approaches.
%We aim to identify the most beneficial way to leverage these objectives through quantitative analysis. 
Experimental results aim to assess how supervised contrastive training can aid the model's ability to classify accurately while inducing meaningful separations between linguistic groups in the embedding space.
Our experiments show that utilizing the contrastive objective as an auxiliary loss during fine-tuning leads to the largest improvements compared to other settings. % the other methods tested.
% We introduce and evaluate different supervised contrastive learning objectives as a pre-training step and additional fine-tuning stage to further improve performance. 
% Our experiments demonstrate these objectives enhance classification accuracy and, most importantly, aid in forming meaningful linguistic groupings within the learned representations. 

% Our main contributions are: (i) We present the first effort at classifying Italian language varieties from speech data alone. (ii) We explore using contrastive learning techniques to improve the model's ability to accurately classify region language varieties while inducing clear separations between linguistic groups in the embedded feature space. (iii) We thoroughly analyze how well models trained with various contrastive objectives capture relationships between data points from the same vs. different Italian regions. We seek to shed light on each model's strengths and limitations in differentiating subtler nuances between certain challenging regions. 
The main contributions of this work are threefold. 
First, it is the first effort at classifying Italian language varieties solely from speech data.
Second, it explores the use of contrastive learning techniques to improve model accuracy in region identification, while also inducing clearer separations between linguistic groups in the embedded feature space. 
Third, it provides an in-depth analysis of how well models trained with different contrastive objectives capture relationships between data points from the same versus different Italian regions. 
The goal is to shed light on the strengths and limitations of the data and the task itself in differentiating subtler nuances between certain challenging regions.

%Advancing the ability to accurately differentiate language varieties has promising applications. 
%For example, it could help documentation efforts for endangered forms by facilitating identification. Also, a more precise classification of locations from language use would allow public and private services to better support cultural preservation initiatives. Finally, location-targeted personalized advertising could be used to better engage the audience.
Advancing the ability to accurately differentiate language varieties may enhance language understanding tasks by exploiting the knowledge of local language varieties.
It also has promising educational and cultural applications.
For example, games or language learning tools could teach aspects of different languages through speech-based interaction and feedback. 
Automatic speech-based region recognition could also help document and analyze lesser-known regional varieties facing endangerment. 
Finally, location-targeted personalized advertising could be used to better engage the audience using familiar local words and idioms to build trust and understanding.

\section{Related Work}
\label{sec:rw}
Research on fine-grained classification of language varieties has advanced in recent years. 
Notable developments include improved data collection and modeling approaches within the speech and natural language processing domains.

\vspace{2mm}
\noindent \textbf{NLP approaches to model language varieties.}
Automatic region classification aims explicitly at predicting the region of origin for linguistic samples based on their textual or linguistic content~\citep[\textit{inter alia}]{gaman-etal-2020-report, dunn-wong-2022-stability}.
This task differs from geolocation~\cite{rahimi-etal-2017-continuous}, which seeks to directly pinpoint the exact coordinates where a text sample was recorded %or generated 
rather than classifying the associated region based on linguistic attributes. %  indicative of a speaker's place of origin. 
It also differs from geo-characterization~\cite{adams2018crowdsourcing}, which focuses on predicting descriptive attributes about a location rather than classifying the region associated with a speaker's linguistic content.

Our primary focus is on using deep learning techniques to detect language variety from speech audio data, however, insights from dialectometry, as discussed by~\citet{Goebl1, Goebl2, geo_diff_tuscany}, suggest the possibility of quantifying the similarity between language varieties.
While progress has been made in NLP and speech technologies for major languages, work explicitly tailored for Italy's language varieties is still relatively limited. 
Recent advancements in Italian-specific NLP models, including both sequence-to-sequence models tailored for Italian \citep{IT5,BARTIT} and adaptations of decoder-only language models \citep{llamantino, camoscio}, have yielded promising results across various tasks. 
However, most NLP research still adopts a monolithic view of Italian as Standard Italian alone, without representation of local languages and regional varieties that characterize the full linguistic landscape~\cite{ramponi2022nlp}.

Additional work tailored to local languages and sociolinguistic factors could advance the field toward solutions that better reflect the full spectrum of language varieties spoken in diverse communities across Italy.
Pioneering efforts in this direction include the DiatopIt corpus~\cite{diatopit}, which stands as the first work focusing on collecting data including diatopic variation beyond Standard Italian %through extensive data collection, 
as well as the related GeoLingIT shared task~\cite{ramponi2023geolingit}, which aimed at identifying the geographic origin of tweets based on their linguistic content.
Building upon this foundation, recent studies by \citet{gallipoli2023dante} and \citet{koudounas2023barhotti} explored novel multi-task learning strategies for enhancing a textual model ability to discriminate between Italian language varieties and solve geolocation challenges.

% The shared task proposed by \citet{ramponi2023geolingit} was among the first efforts to automatically model regional variation in Italian language varieties using a large Twitter corpus.
% Additional work tailored to local languages and sociolinguistic factors has the potential to advance the field towards solutions truly reflective of Italy's full spectrum of language codes as spoken in diverse communities nationwide~\cite{gallipoli2023dante, koudounas2023barhotti}.
% \cite{gallipoli2023dante} proposed a multi-task pre-training approach operating at both the token and sentence levels to enhance performance for related tasks.
% \cite{koudounas2023barhotti} leveraged contrastive learning objectives to enhance representations and final performance, while also employing multi-task learning.
% One pioneering effort in this direction is the DiatopIt corpus~\cite{diatopit}, which stands as the first work focusing on depicting diatopic variation beyond Standard Italian. 

Recent work has also aimed at developing speech understanding systems for Italian, such as EMOVO~\cite{costantini2014emovo} targeting emotional speech, or IDEA~\cite{IDEA} modeling dysarthric speech using isolated words. 
However, these datasets offer limited domain coverage or lack information on speakers' regional origins.
% The AlmaWave-SLU corpus~\cite{Almawave-SLU} translates English utterances rather than capturing native Italian audio. 
ITALIC~\cite{italic} is the largest Italian speech dataset for intent classification. However, while it collects information on speakers' origins, the recordings are in Standard Italian rather than regional languages.

%\subsection{Identifying language varieties from speech}
%\label{subsec:sli}
\vspace{2mm}
\noindent \textbf{Identifying language varieties from speech.}
The task of automatically identifying the language of a speech audio recording is commonly known as spoken language recognition. %, or SLI for short. 
Large-scale evaluations like the Language Recognition Evaluation (LRE) campaigns have addressed spoken language identification at a broader scale, evaluating systems on a wide range of languages from around the world~\cite{lre_2022, lre_17}.
Recent research has shown that convolutional neural networks and transformer-based models can both achieve high accuracy for language recognition tasks. 
CNN architectures have been shown to reach state-of-the-art performance even with limited training data~\cite{sarni23_interspeech}. 
On the other hand, transformers leverage self-supervised pre-training to capture robust linguistic patterns and learn broadly generalizable representations~\cite{alumae23_interspeech}.
%Recent research has demonstrated that m
Models employing the Wav2Vec 2.0~\cite{wav2vec2} architecture can inherently capture language-discriminative information in their lower layers~\cite{accidental_learners}. 
These models can classify unseen languages and adapt to new conditions without further training~\cite{efficient_sli_ssl}, suggesting such capabilities may also enable the models to distinguish closely related languages when provided with specific supervision.
The latent, discretized representations learned by these models are shared across languages~\cite{sli_discrete_representations}, allowing for effective language identification.
Several studies have already extended this research line to the finer identification of linguistic variations across geographical regions.
Developing modeling approaches that can distinguish local varieties requires targeted corpora representing regional linguistic variation.
\citet{finnish_dataset} introduced the \textit{Lahjoita puhetta} corpus containing diverse samples of colloquial Finnish speech reflecting different sociolinguistic factors and dialects. 
Similarly, \citet{romanian_dialect_dataset} contributed to the RoDia dataset containing Romanian dialectal speech from various regions. 
These resources open opportunities for designing specific models to analyze curated speech corpora.
\citet{german_speech_dialect} demonstrated the effectiveness of combining selected spectral features with Gaussian mixture models for dialect discrimination and classification tasks.
\citet{finnish_dialect_identification} assessed both textual and multi-modal classifiers for Finnish dialects, highlighting the importance of leveraging the audio modality to discriminate nuanced dialectal differences. 
Additionally, \citet{northsami_dialect} showcased self-supervised speech models' ability to distinguish between four variants of North Sámi.
Together, these findings indicate the potential of pre-trained speech models for fine-grained language and variety discrimination tasks.

\section{Methodology}
\label{sec:method}

To address the proposed task, we leverage multilingual pre-trained models trained on large datasets to learn general representations.
We also investigate the use of contrastive learning objectives to enhance the fine-tuning process and better separate the embeddings of different regions. %better separate the embeddings by region. 
% Those objectives can be either applied as an additional supervised pre-training step before the final fine-tuning or directly integrated into the fine-tuning stage as an extra training criterion to optimize the model.
This section provides an in-depth examination of these two key aspects: % of the proposed methodology.
Section~\ref{subsec:contrastive_expl} describes the contrastive learning objectives explored, while Section~\ref{subsec:finetuning} explains how fine-tuning is performed to adapt the pre-trained models to the region identification task.
% This section provides an in-depth examination of these two key aspects of the proposed methodology, namely the contrastive learning objectives (Section~\ref{subsec:contrastive_expl}) and the fine-tuning procedure followed for adapting the pre-trained models to the region identification task (Section~\ref{subsec:finetuning}).

\subsection{Contrastive Learning Objectives}
\label{subsec:contrastive_expl}

Contrastive learning is a representation learning method that focuses on acquiring knowledge by comparing and contrasting positive and negative examples.
In a nutshell, the model learns class-discriminative representations by maximizing the similarity between representations of positive examples and minimizing similarity for negatives.
We evaluate several supervised contrastive loss functions to improve the model's ability to learn discriminative representations.
Specifically, we examine supervised contrastive loss (SC), triplet margin loss (TM), and multi-similarity loss (MS) as additional training objectives.

\vspace{2mm}
\noindent \textbf{Supervised contrastive loss (SC).} 
Defined as a direct optimization of embedding similarities between positive and negative pairs, the supervised contrastive loss~\cite{khosla2020supervised} aims to maximize agreement between samples from the same region while minimizing agreement between samples from different regions. 
%Given a batch of audio samples $\mathbf{x}_i$ and corresponding region labels $\mathbf{y}_i$, 
Let $f(\mathbf{x}_i)$ be an encoder network that generates an embedding $\mathbf{z}_i = f(\mathbf{x}_i)$ for each audio sample $\mathbf{x}_i \in \mathcal{X}$. 
The SC loss is then defined as:
\begin{equation}
    \mathcal{L}_{SC} = - \sum_{i \in \mathcal{I}} \dfrac{1}{|\mathcal{P}_i|} \sum_{p \in \mathcal{P}_i} \log \dfrac{\exp(\mathbf{z}_i \cdot \mathbf{z}_p / \tau)}{\sum_{n \in \mathcal{N}_i} \exp(\mathbf{z}_i \cdot \mathbf{z}_n / \tau)}
\end{equation}
Where $\mathcal{I}$ is the set of all samples in a given batch, $\mathcal{P}_i$ is the set of positive samples for sample $i$, $\mathcal{N}_i$ is the set of all negative samples for sample $i$, and $\tau$ is a tunable parameter. 
The positive samples are defined as all samples in the same training batch having the same label (i.e., the same region of origin in our case), while the negative samples are all samples in the same batch having a different label.
The temperature $\tau$ is a scaling fudge factor 
%hyperparameter 
we set to $0.1$, following the default value used in the original paper~\cite{khosla2020supervised}.

\vspace{2mm}
\noindent \textbf{Triplet margin loss (TM).}
Similar to supervised contrastive loss, the triplet margin loss~\cite{triplet_margin_loss} aims to maximize similarities between embeddings of positive samples and minimize similarities between negatives. 
However, it operates using triplets of samples during training rather than pairs.
Given a triplet consisting of an anchor sample $\mathbf{x}_a$, a positive sample $\mathbf{x}_p$ 
and a negative sample $\mathbf{x}_n$, the TM loss is defined as:
\begin{equation}
    \mathcal{L}_{TM} = \max(0, d(\mathbf{z}_a, \mathbf{z}_p) - d(\mathbf{z}_a, \mathbf{z}_n) + \mu)
\end{equation}
Where $d(\cdot, \cdot)$ is a distance function measuring the distance between embeddings; we use the L2 distance here. 
The margin parameter $\mu$ regulates the desired separation between positive and negative samples, with a higher value resulting in greater separation. In our experiments, we set $\mu=0.05$.
Triplets are generated using all combinations of samples within each batch, where the anchor and its positives belong to the same class, and the negatives belong to a different class. 
Intuitively, this loss function aims to minimize distances between positive embeddings while simultaneously enforcing a minimum margin between negative embeddings relative to the anchor.

\vspace{2mm}
\noindent \textbf{Multi-similarity loss (MS).} The multi-similarity loss~\cite{wang2019multi} is a pair-based contrastive loss that specifically models pair selection and weighting to pick the most informative pairs for training. 
It optimizes pair selection and weighting by considering the similarity of a given sample to: (i) itself, (ii) other samples from the same class, and (iii) samples from different classes.
Once pairs are selected, the loss is computed as the sum of positive and negative terms:
\begin{equation}
    \begin{aligned}
        \mathcal{L}_{MS} = \dfrac{1}{m} \sum_{i=1}^m  
        \dfrac{1}{\alpha} \log [ 1 + \sum_{p \in \mathcal{P}_i} e^{-\alpha (S_{ip} - \lambda)} ] \\
        + \dfrac{1}{\beta} \log [ 1 + \sum_{n \in \mathcal{N}_i} e^{\beta (S_{in} - \lambda)} ] 
    \end{aligned}
\end{equation}
Where $m$ is the batch size, $\mathcal{P}_i$ and $\mathcal{N}_i$ denote the sets of positive and negative samples for the anchor $i$, $S_{ip}$ and $S_{in}$ are the similarities between $i$ and its positive/negative pairs, and $\alpha, \beta, \lambda$ control pair weighting.
The multi-similarity loss aims to select the most informative sample pairs during training by adaptively weighting pairs based on their similarities, prioritizing the most discriminative contrasts.

\vspace{2mm}
\noindent
We investigate the use of each contrastive objective in three different settings: (i) the contrastive loss is used as the sole objective function to optimize the model parameters during an additional pre-training phase, (ii) the contrastive loss is added as an extra term to the overall loss function during the fine-tuning stage, and (iii) a combination where the model first undergoes pre-training where contrastive loss is minimized, then during the fine-tuning stage both contrastive and task-specific classification loss terms are jointly optimized.

\subsection{Model fine-tuning}
\label{subsec:finetuning}

To adapt the pre-trained models for the region identification task, we fine-tune them on VIVALDI dataset, which is better described in Section \ref{sec:data}. % on the labeled audio collection.
Most of the models tested in this analysis generate a high-level vector representation for each audio frame (i.e., typically 20 ms) after applying a contextualized encoder to raw audio.
We perform average pooling over all frame embeddings to obtain a single embedding representing the full audio recording.
Specifically, for a given recording composed of $T$ audio frames with corresponding embeddings ${e_1, e_2, ..., e_T}$, we obtain: $e = \frac{1}{T} \sum_t e_t$, where $e$ is the overall recording embedding. 
Taking the average in this manner aggregates the linguistic information captured across all frames into a single fixed-dimensional vector for the recording.
Other methods of obtaining a recording-level representation from frame embeddings (e.g., attention pooling) could also be explored. 
Investigating alternative pooling strategies is outside the scope of the current work but may be considered in future explorations seeking improved performance.

%The model is trained end-to-end to optimize the classification objective during fine-tuning. 
Each model is trained end-to-end on the labeled audio collection by minimizing the cross-entropy loss between the model's region predictions and reference labels, directly optimizing its weights for the region identification task. 
Optionally, a contrastive learning objective (see Section~\ref{subsec:contrastive_expl}) may also be included to simultaneously address the classification task and enrich the quality of data representation.

\section{Experiments}
\label{sec:experiments}

To evaluate the proposed approach, we conduct a series of experiments to separately analyze the performance of different pre-trained models and the impact of the contrastive learning objectives.

\subsection{VIVALDI data collection}
\label{sec:data}

\begin{table}[]
\resizebox{\columnwidth}{!}{%
\begin{tabular}{@{}ccccc@{}}
\toprule
Split & \# Samples & \# Minutes & \# Cities & Avg \# per city \\ \midrule
Train & 81279      & 2253.88    & 237       & 344.40          \\
Val   & 8242       & 228.98     & 24        & 343.42          \\
Test  & 8241       & 227.52     & 24        & 343.38          \\ \bottomrule
\end{tabular}%
}
\caption{VIVALDI dataset statistics.}
\label{tab:split_stats}
\end{table}

The VIVALDI dataset~\cite{vivaldi}, short for "Vivaio Acustico delle Lingue e dei Dialetti d’Italia", is a comprehensive collection of spoken utterances from nearly all regions of Italy \cite{vivaldi_sicilia, vivaldi_sardegna, vivaldi_umbria}.
Italy comprises 20 administrative regions, and VIVALDI covers recordings from 19 of them, including cities from all regions except \textit{Marche}.
% This extensive collection captures the linguistic diversity found throughout the country.
While most regions contain samples from 3 or more cities, two regions, Lazio and Campania, are each represented by data from a single city only.
Despite variations in city-level coverage between regions, %, this collection captures the rich linguistic diversity found throughout Italy. 
to the best of our knowledge, this is the only data collection including extensive speech recordings in local languages from a wide range of locations across the country, allowing investigation at an unprecedented scale.

% The dataset consists of a specific set of sentences, averaging approximately 343 per city, spoken in various areas across Italy with variations and potentially large discrepancies in the number of cities included from each region. 
The VIVALDI collection comprises recordings of a specific set of sentences, averaging approximately 343 repetitions per city. 
Each sample included in the collection is geolocated, providing information about the specific city where it was recorded.
In most cases, each city is represented by recordings from a single speaker, with the sole exception being \textit{Aidone} in \textit{Sicily} region where two different speakers were sampled.
% This geographical context is crucial for our analysis of regional linguistic variations.
Unfortunately, due to the unavailability of speaker-related information, conducting specific analyses or differentiating results based on demographic factors is currently unfeasible. Nonetheless, the dataset's geographical context remains pivotal for our analysis of regional linguistic variations.

\vspace{2mm}
\noindent \textbf{Dataset statistics.} 
Table~\ref{tab:split_stats} summarizes the key dataset statistics, including the number of samples, total minutes of audio, cities covered, and the average number of samples per city for the three splits: training, validation, and test.
The dataset isn't provided as a single unified collection on the official website, and to the best of our knowledge, no predefined splits are available. Consequently, we design these splits so that each city, and each speaker accordingly, is allocated to only one of the splits (train, validation, or test), to ensure that models can not rely on identifying individuals to determine a sample's region.
%The dataset splits are designed so that each city, and consequently each speaker, is allocated to only one of the splits (train, validation, or test). 
%The dataset isn't provided as a single unified collection, and to the best of our knowledge, no predefined splits are available. 
%Consequently, we design these splits to ensure that models can not rely on identifying individuals to determine a sample's region.
Instead, the task is centered on classifying each recording based on its linguistic features.
While an 80/10/10 train/validation/test split was initially intended, we applied some constraints to enforce that each split contains samples from at least one city of each region. 
Applying this constraint, regions with data from fewer than three cities were excluded from the dataset, resulting in a final dataset representing %only 
17 out of the 20 Italian regions.
This split makes the classification task very challenging as models must generalize to unseen cities within a region that may also exhibit local linguistic variations. 
% Unfortunately, due to the unavailability of speaker-related information, we cannot conduct specific analyses or differentiate results based on demographic factors.

\subsection{Experimental settings}
\label{subsec:exp_settings}

%The experiments are conducted using the VIVALDI dataset~\cite{vivaldi} described in Section~\ref{sec:data}.
As evaluation metrics, we consider accuracy and macro F1 score. 
Accuracy is defined as the percentage of samples correctly classified, while macro F1 score represents the unweighted average of the F1 scores for each class. 
Given the imbalanced nature of the dataset, the macro F1 score provides a more reliable metric for evaluating the model's performance. 
We report the mean and standard deviation of results across three independent runs for each experiment.

\subsection{Model selection}
\label{subsec:model_selection}

We explore the following speech models to assess their performance %of various pre-trained models 
on the VIVALDI dataset.

\vspace{1mm}
\noindent\texttt{WavLM}%\footnote{\url{https://huggingface.co/microsoft/wavlm-large}}
~\cite{wavlm}: a model trained on 84,000 hours of audio from Libri-Light~\cite{librilight}, GigaSpeech~\cite{gigaspeech}, and VoxPopuli~\cite{voxpopuli} corpora. As the model was trained primarily on English data, it includes only limited multilingual capabilities.

\vspace{1mm}
\noindent\texttt{XLSR-53}
~\cite{xlsr53}: a multilingual model trained on 53 languages from CommonVoice~\cite{commonvoice}, Multilingual LibriSpeech~\cite{mls}, and BABEL~\cite{babel} datasets.

\vspace{1mm}
\noindent\texttt{XLSR-128}
~\cite{xlsr128}: a multilingual model trained on 128 languages including the same datasets as \texttt{XLSR-53} in addition to VoxPopuli~\cite{voxpopuli} and VoxLingua107~\cite{voxlingua}.

\vspace{1mm}
\noindent\texttt{ECAPA}
~\cite{ecapa}: a CNN-based spoken language identification model pre-trained on 107 languages from VoxLingua107~\cite{voxlingua} dataset using SpeechBrain toolkit~\cite{speechbrain}.

% \begin{itemize}
%     \item \texttt{WavLM}%\footnote{\url{https://huggingface.co/microsoft/wavlm-large}}
%     ~\cite{wavlm}: a primarily English model trained on 84,000 hours of audio from Libri-Light~\cite{librilight}, GigaSpeech~\cite{gigaspeech}, and VoxPopuli~\cite{voxpopuli} corpora. As the model was trained primarily on English data, it includes only limited multilingual capabilities.
%     \item \texttt{XLSR-53}%\footnote{\url{https://huggingface.co/facebook/wav2vec2-large-xlsr-53}}
%     ~\cite{xlsr53}: a multilingual model trained on 53 languages from CommonVoice~\cite{commonvoice}, Multilingual LibriSpeech~\cite{mls}, and BABEL~\cite{babel} datasets.
%     \item \texttt{XLSR-128}%\footnote{\url{https://huggingface.co/facebook/wav2vec2-xls-r-300m}}
%     ~\cite{xlsr128}: a multilingual model trained on 128 languages including the same datasets as \texttt{XLSR-53} in addition to VoxPopuli~\cite{voxpopuli} and VoxLingua~\cite{voxlingua} datasets.
%     \item \texttt{ECAPA}%\footnote{    \url{https://huggingface.co/speechbrain/lang-id-voxlingua107-ecapa}}
%     ~\cite{ecapa}: a spoken language identification model pre-trained on 107 languages from VoxLingua107~\cite{voxlingua} dataset using SpeechBrain toolkit~\cite{speechbrain}.
% \end{itemize}

\vspace{1mm}
% For \texttt{XLSR-53} %\footnote{\url{https://huggingface.co/jonatasgrosman/wav2vec2-large-xlsr-53-italian}}
% and \texttt{XLSR-128} %\footnote{\url{https://huggingface.co/dbdmg/wav2vec2-xls-r-300m-italian}} 
% models we also evaluate Italian fine-tuned versions of the models, that have been trained for Italian ASR tasks on the CommonVoice~\cite{commonvoice} dataset.
For the \texttt{XLSR-53} and \texttt{XLSR-128} models, we also evaluate versions that have been additionally fine-tuned for Italian ASR tasks on the CommonVoice~\cite{commonvoice} dataset (i.e., \texttt{XLSR-53-ITA} and \texttt{XLSR-128-ITA}, respectively).
% Those models should be able to better capture the linguistic nuances of Italian speech. 
% Given the limited amount of Italian data available, we expect them to perform better than the multilingual models. 
Those models may better capture the linguistic variation across Italy's regional forms.

All models are fine-tuned using the same training and validation splits described in Section~\ref{sec:data} and following the same training procedure.
To this end, we use the AdamW optimizer with 10\% of warm-up steps and a linear learning rate decay schedule.
The maximum learning rate is set to $10^{-4}$, and the weight decay is set to $10^{-2}$.
We train each model with a batch size of 32 for a maximum of 10 epochs, selecting the best model based on the validation loss.
Specific details about the models used in this analysis, together with the corresponding hyperparameters and the code to reproduce the experiments, are available on the project's repository\footnote{\label{repo_url}\url{https://github.com/MorenoLaQuatra/SALVI}}.

\subsection{Quantitative Results}

\noindent \textbf{Standard fine-tuning.} %The models described in Section~\ref{subsec:model_selection} are evaluated on the city-level split of data presented in Section~\ref{sec:data}. 
Table~\ref{table:models} reports the mean and standard deviation of accuracy and macro F1 scores for models trained using standard fine-tuning for the downstream task of region classification.
% These metrics were calculated from three independent runs of standard fine-tuning for the downstream task of region classification.
Overall, the best-performing model is \texttt{XLSR-53-ITA}, %further fine-tuned for Italian ASR
achieving a macro F1 score close to 50\%. 
Interestingly, the raw version of \texttt{XLSR-53} without Italian fine-tuning outperforms even the \texttt{XLSR-128-ITA} model and its pre-trained version without additional fine-tuning. 
This is likely because while the model architectures are identical, \texttt{XLSR-128} covers a much higher number of languages in its pre-training and thus may lack the capacity to capture fine-grained linguistic variations.
\texttt{WavLM} does not achieve good results, because it was only partially pre-trained on multilingual data, which is insufficient for capturing the specific nuances information needed to solve the non-trivial region classification problem on the VIVALDI data collection. 
\texttt{ECAPA} is the only model leveraging a CNN-based architecture and is designed for spoken language identification and trained on 107 languages (including Italian). 
It performs  worst among the tested models with a macro F1 score below 20\%.

\begin{table}
\centering
\resizebox{\columnwidth}{!}{%
\begin{tabular}{cccc}
\toprule
\textbf{Model} 
    & \textbf{ITA-FT} 
    & \textbf{Accuracy} 
    & \textbf{F1 Macro} \\
\midrule

% \texttt{ECAPA} 
%     & \xmark
%     & 23.53$\pm$0.39
%     & 18.42$\pm$0.18 \\

\texttt{WavLM-L} 
    & \xmark
    & 53.35$\pm$1.62
    & 43.76$\pm$1.14 \\
    
\texttt{XLSR-53} 
    & \xmark
    & 56.99$\pm$0.61
    & 48.02$\pm$1.13 \\
    
\texttt{XLSR-128} 
    & \xmark
    & 52.85$\pm$2.03
    & 44.95$\pm$2.28 \\ \midrule

\texttt{XLSR-53-ITA} 
    & \cmark
    & \textbf{60.18$\pm$0.55}
    & \textbf{49.84$\pm$0.57} \\

\texttt{XLSR-128-ITA} 
    & \cmark
    & 55.62$\pm$2.24
    & 47.83$\pm$2.33 \\
    
\bottomrule

\end{tabular}
}
\caption{\label{table:models} Model selection. Mean and standard deviation results across three independent runs considering different models with a classic fine-tuning approach. The ITA-FT column indicates whether models were fine-tuned for Italian ASR (\cmark) or not (\xmark). Best results are highlighted in bold.} %The best results are in bold.}
\end{table}

\vspace{2mm}
\noindent \textbf{Contrastive learning.}
We experiment with three different supervised contrastive losses used in three different settings % as pre-training objectives, additional fine-tuning objectives, and in combination, 
as explained in Section~\ref{sec:method}.
Table~\ref{table:results} shows the results of our approach using the previously best-performing model, \texttt{XLSR-53-ITA}. %fine-tuned for Italian ASR. 
We separately indicate the use of standard classification objective during fine-tuning (\textbf{Clf-FT}), contrastive objective as additional fine-tuning task (\textbf{Ctr-FT}), and contrastive objective as additional pre-training step (\textbf{Ctr-PT}).
% Applying multi-similarity and triplet margin losses as additional contrastive objectives helps improve performance on the downstream task in all tested configurations, namely multi-task fine-tuning with classification (\textbf{Clf-FT} in the table) and contrastive (\textbf{Ctr-FT}) objectives, contrastive pre-training (\textbf{Ctr-PT}) followed by standard classification fine-tuning, and contrastive pre-training followed by multi-task fine-tuning. 
Applying multi-similarity and triplet margin losses as additional contrastive objectives helps improve performance on the downstream task in all tested configurations. 
Models trained using multi-similarity objective achieve the best overall results in all three configurations, with the highest macro F1 score of 51.29\% in the multi-task fine-tuning scenario. 
Conversely, applying the supervised contrastive loss consistently results in decreased performance, especially as a pre-training step.
% This suggests that even if both losses contrast sample pairs, the weighting scheme proposed in MS loss allows it to better capture nuanced linguistic variations compared to the standard Supervised Contrastive loss.
The multi-similarity loss prioritizes the most informative contrasts during training by adaptively weighting sample pairs based on their similarities. 
This fine-grained approach allows the model to better learn representations that can capture subtle linguistic variations between regions, compared to the standard supervised contrastive loss, which treats all negative pairs equally.

Regardless of the specific objective, the best results are achieved using the additional contrastive loss for multi-task fine-tuning only. 
% Applying it as a pre-training step followed by either traditional or multi-task fine-tuning does not match this performance, likely because limited data prevents the model from fully capturing the task's complex nuances.
Combining objectives uniquely during fine-tuning leverages their complementary strengths by directly shaping the representations for the downstream task. 
Earlier-stage usage lacks this targeted optimization and results to slightly worsened final performance on the identification task.

\begin{table}[]
\resizebox{\columnwidth}{!}{%
\begin{tabular}{@{}ccccc@{}}
\toprule
\textbf{Ctr-PT} 
    & \textbf{Ctr-FT} 
    & \textbf{Clf-FT} 
    & \textbf{Accuracy} 
    & \textbf{F1 Macro} \\ 
\midrule

% Original FT
\xmark                          % Ctr-PT
    & \xmark                    % Ctr-FT
    & \cmark                    % Clf-FT
    & 60.18$\pm$0.55        % Accuracy
    & 49.84$\pm$0.57 \\     % F1 Macro
\midrule

% Supervised Contrastive Loss
\multicolumn{5}{c}{Supervised Contrastive Loss}  \\ \midrule
\xmark                          % Ctr-PT
    & \cmark                    % Ctr-FT
    & \cmark                    % Clf-FT
    & 59.02$\pm$1.26            % Accuracy
    & 49.31$\pm$1.33 \\         % F1 Macro
\cmark                      % Ctr-PT
    & \xmark                    % Ctr-FT
    & \cmark                    % Clf-FT
    & 58.98$\pm$0.67            % Accuracy
    & 48.82$\pm$0.44 \\         % F1 Macro
\cmark                      % Ctr-PT
    & \cmark                    % Ctr-FT
    & \cmark                    % Clf-FT
    & 57.46$\pm$2.10            % Accuracy
    & 47.53$\pm$1.44 \\         % F1 Macro    
\midrule

% Triplet Loss
\multicolumn{5}{c}{Triplet Margin Loss}  \\ \midrule
\xmark                          % Ctr-PT
    & \cmark                    % Ctr-FT
    & \cmark                    % Clf-FT
    & 60.23$\pm$1.53            % Accuracy
    & 50.68$\pm$1.71 \\         % F1 Macro
\cmark                          % Ctr-PT
    & \xmark                    % Ctr-FT
    & \cmark                    % Clf-FT
    & 59.87$\pm$0.58            % Accuracy
    & 50.56$\pm$0.50 \\         % F1 Macro
\cmark                          % Ctr-PT
    & \cmark                    % Ctr-FT
    & \cmark                    % Clf-FT
    & 58.19$\pm$0.64            % Accuracy
    & 49.92$\pm$1.14 \\         % F1 Macro
\midrule

% Multi-Similarity Loss
\multicolumn{5}{c}{Multi-Similarity Loss} \\ \midrule
\xmark                          % Ctr-PT
    & \cmark                    % Ctr-FT
    & \cmark                    % Clf-FT
    & \textbf{60.49$\pm$0.88}   % Accuracy
    & \textbf{51.29$\pm$1.36} \\% F1 Macro
\cmark                      % Ctr-PT
    & \xmark                    % Ctr-FT
    & \cmark                    % Clf-FT
    & 58.92$\pm$1.35            % Accuracy
    & 51.07$\pm$0.61 \\         % F1 Macro  
\cmark                      % Ctr-PT
    & \cmark                    % Ctr-FT
    & \cmark                    % Clf-FT
    & 59.86$\pm$0.83            % Accuracy
    & 50.98$\pm$0.35  \\        % F1 Macro  
\bottomrule

\end{tabular}
}
\caption{\label{table:results} Mean and standard deviation results across three runs considering pre-training (\textbf{Ctr-PT}) and fine-tuning approaches (\textbf{Ctr-FT} and \textbf{Clf-FT}) with different contrastive losses over the best-performing model. The best results are in bold. % \texttt{XLSR-53} pre-trained for Italian ASR.
}
\end{table}

\subsection{Region Correlation Analysis}

\begin{figure*}
    \begin{subfigure}{0.48\linewidth}
        \centering
        \includegraphics[width=\linewidth]{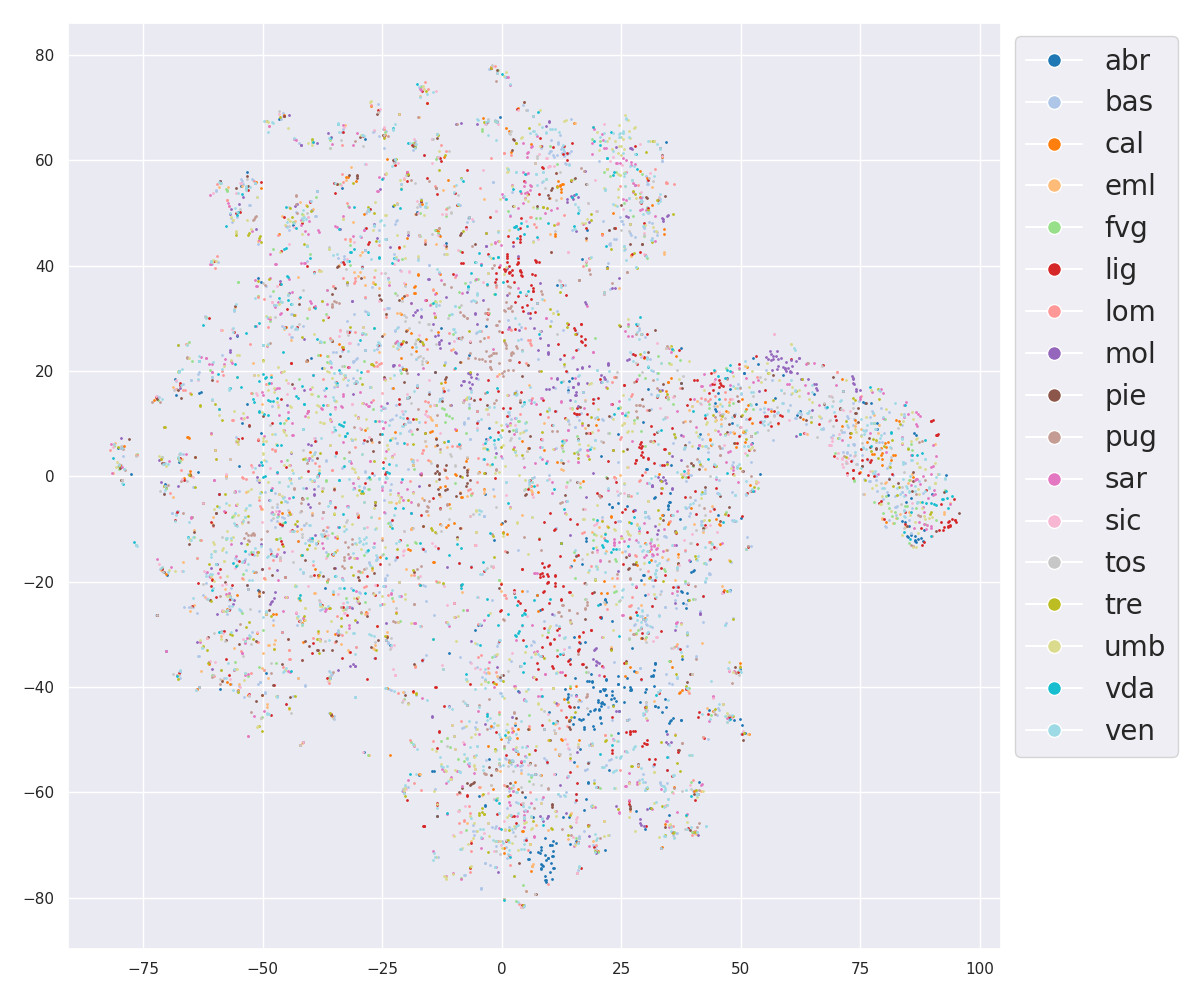}
        \caption{Original model.}
         \label{fig:tsne-original}
    \end{subfigure}
    \hfill
    \begin{subfigure}{0.48\linewidth}
        \centering
        \includegraphics[width=\linewidth]{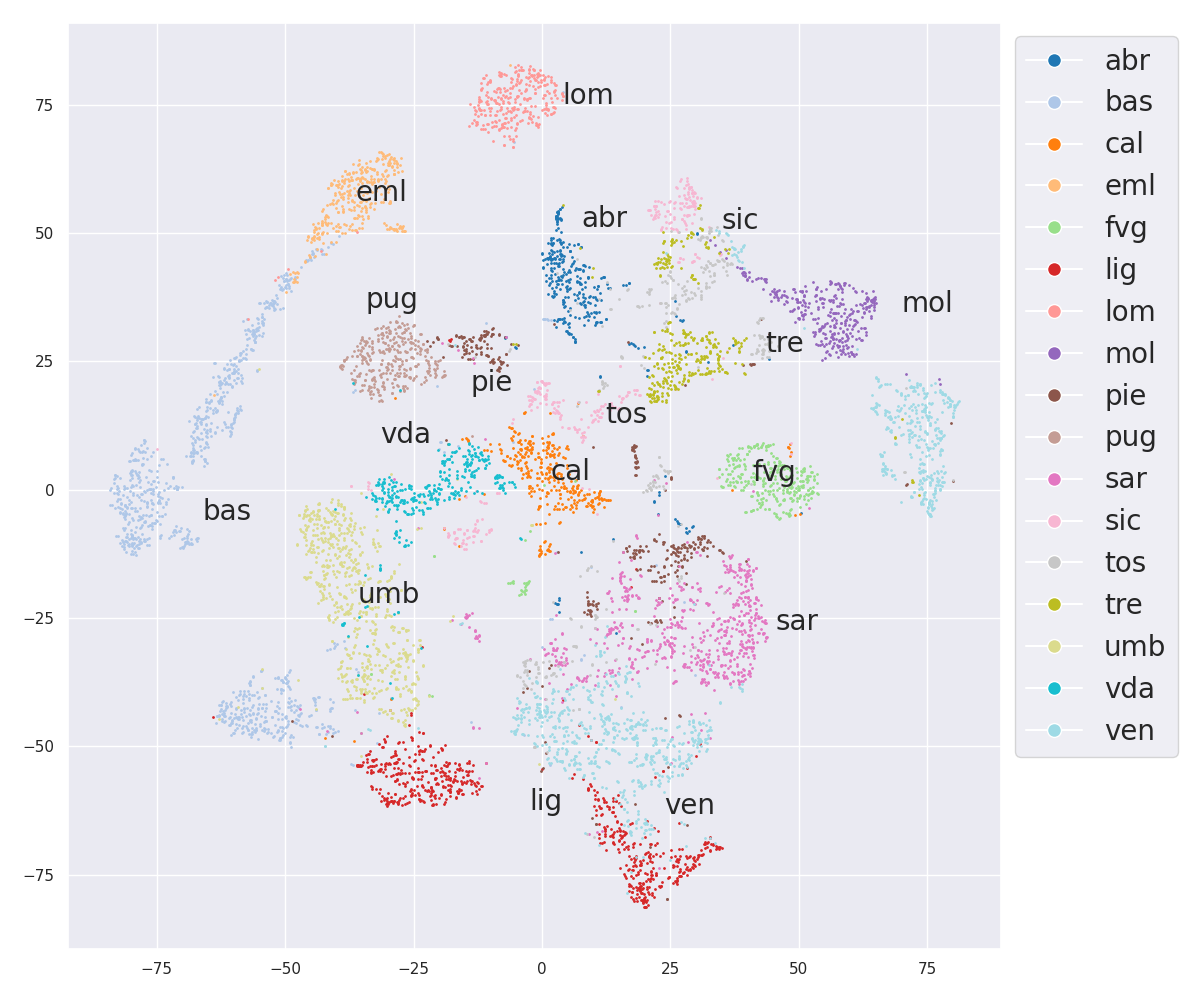}
        \caption{Supervised Contrastive loss.}
         \label{fig:tsne-sc}
    \end{subfigure}
    \begin{subfigure}{0.48\linewidth}
        \centering
        \includegraphics[width=\linewidth]{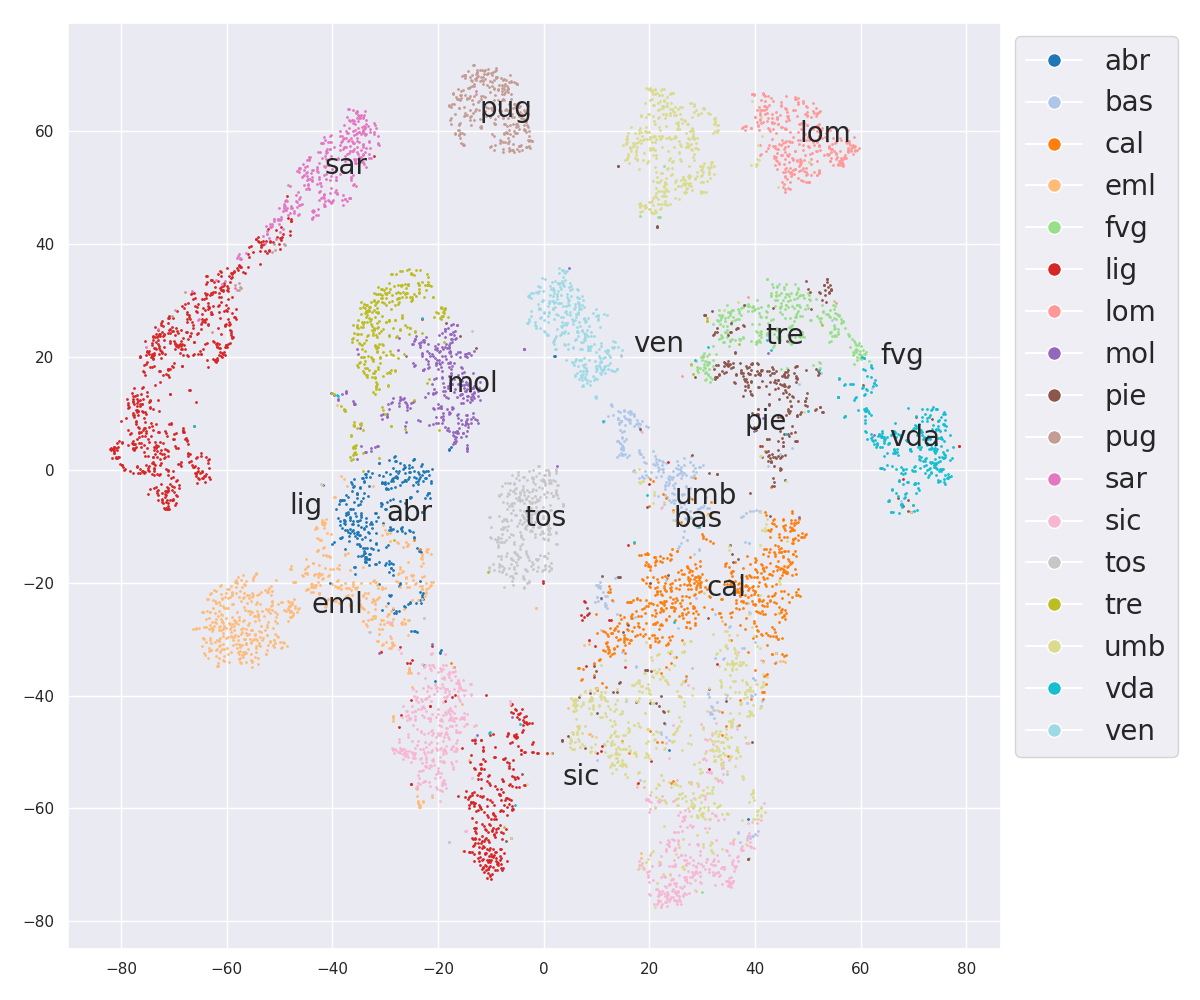}
        \caption{Triplet loss.}
        \label{fig:tsne-tm}
    \end{subfigure}
    \hfill
    \begin{subfigure}{0.48\linewidth}
        \centering
        \includegraphics[width=\linewidth]{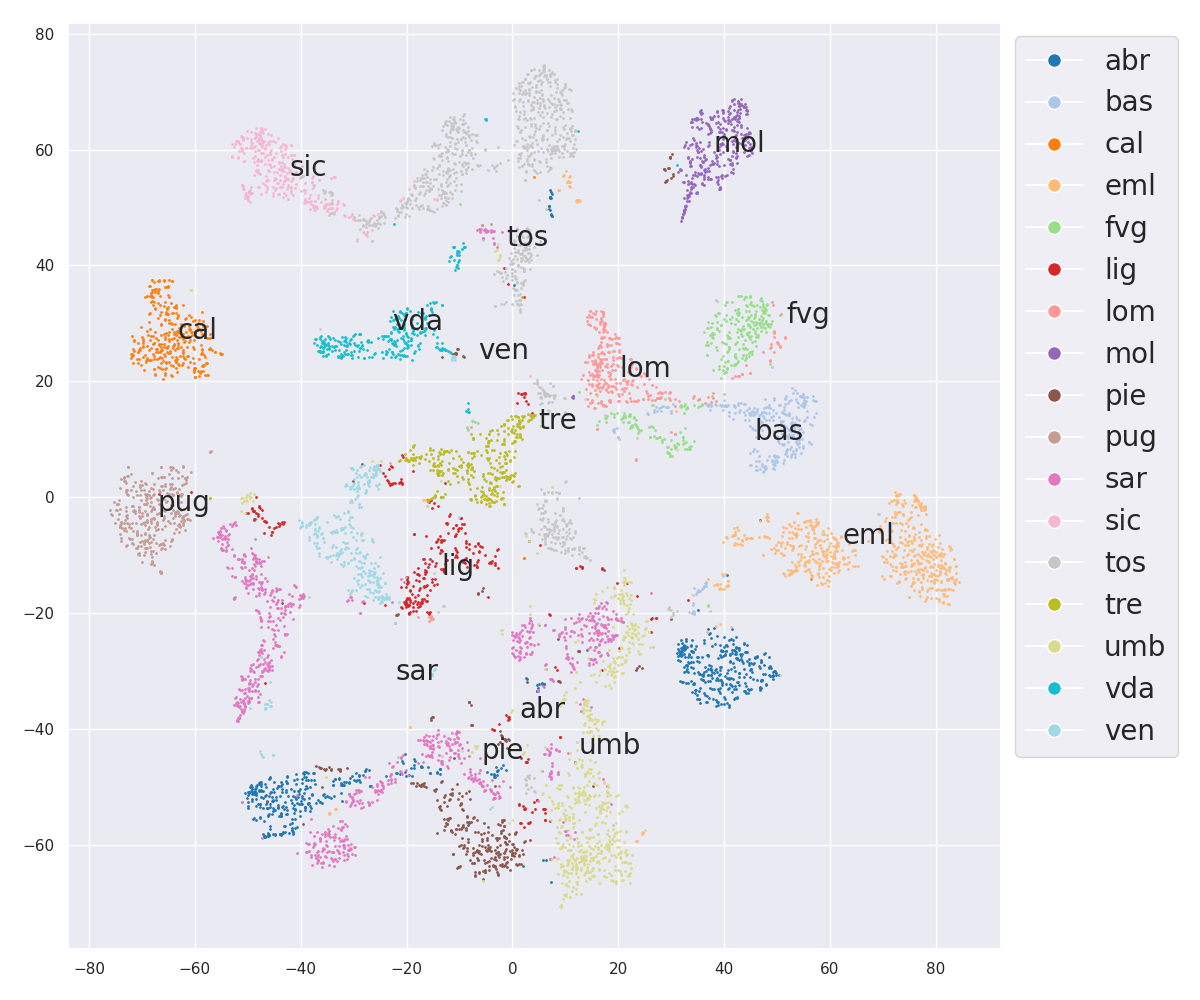}
        \caption{Multi-similarity loss.}
         \label{fig:tsne-ms}
    \end{subfigure}
    \caption{t-SNE visualization of the original \texttt{XLSR-53-ITA} model (a) and the corresponding pre-trained versions with the three different contrastive learning objectives: supervised contrastive loss (b), triplet-margin loss (c), and multi-similarity loss (d).}
    \label{fig:tsne-pt}
\end{figure*}

\noindent \textbf{Pre-training objectives.} 
We analyze the structure of models' high-dimensional feature spaces to gain additional insights into how effectively the pre-training objectives capture relationships between data points. %from the same or different Italian regions. 
Specifically, we apply t-Distributed Stochastic Neighbor Embedding (t-SNE)~\cite{van2008visualizing}, a widely used nonlinear dimensionality reduction technique, to project the embeddings into a two-dimensional space for visualization while attempting to preserve the local geometric structure of the original data.
The resulting projection examines how each model distributes data points relative to their ground-truth region labels in the lower-dimensional space. 
An ideal clustering would show clear separations between different regions, with points from the same region closely located together. 
This analysis is performed using only test set data and aims to characterize each method's capacity to learn discriminative representations that properly distinguish between regions.  

Figure~\ref{fig:tsne-pt} shows t-SNE projections of the embeddings from the original \texttt{XLSR-53-ITA} model (Fig.~\ref{fig:tsne-original}) and the same model pre-trained with different contrastive objectives: supervised contrastive loss (Fig.~\ref{fig:tsne-sc}), triplet margin loss (Fig.~\ref{fig:tsne-tm}), and multi-similarity loss (Fig.~\ref{fig:tsne-ms}).
The data points are color-coded by their ground-truth region labels.

In Figure~\ref{fig:tsne-original}, the representations show no clear clustering, as expected, since the model was not trained for this specific task and does not encode \textit{a priori} the differences between language varieties. 
On the other hand, contrastive pre-training facilitates the formation of more distinct clusters across regions. Figure~\ref{fig:tsne-sc} with SC loss exhibits the most overlapping clusters. Figures~\ref{fig:tsne-tm} and~\ref{fig:tsne-ms} with TM and MS losses exhibit more precise separation of clusters corresponding to different regions, with multi-similarity (Fig.~\ref{fig:tsne-ms}) displaying slightly more well-defined clusters. This corroborates the quantitative results shown in Table~\ref{table:results}, where multi-similarity training led to the best performance gains. 
The projections indicate that contrastive pre-training helps to learn embeddings that more distinctly encode relationships between data points, especially with margin-based losses. %from the same versus different Italian regions.

\begin{figure*}
    \begin{subfigure}{0.49\linewidth}
        \centering
        \includegraphics[width=\linewidth]{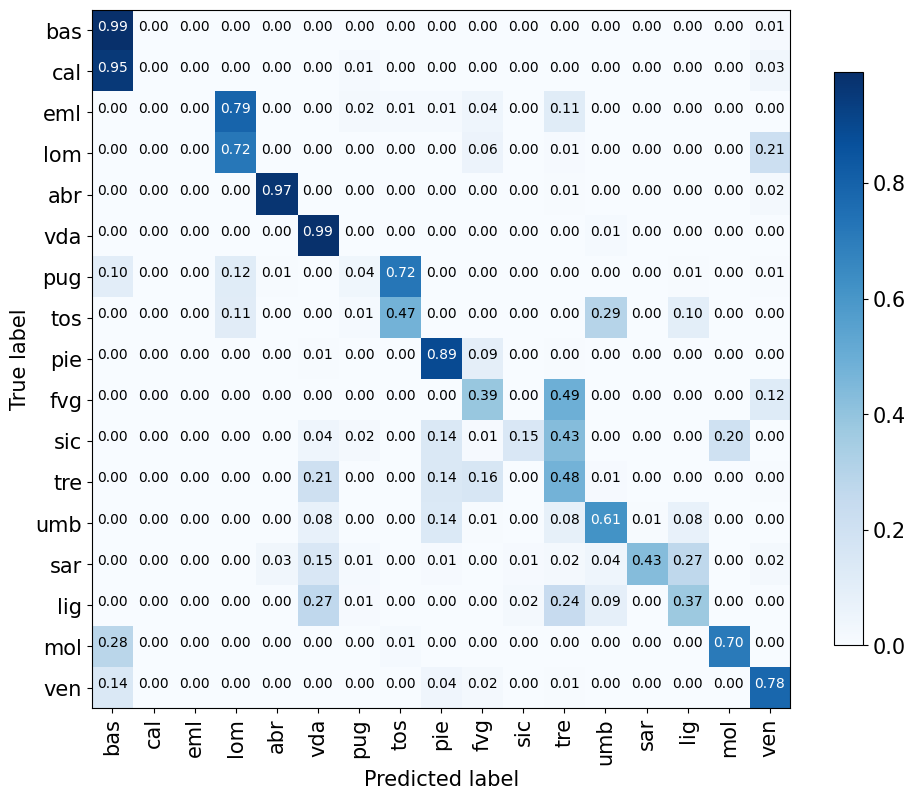}
        \caption{Confusion matrix.}
        \label{fig:best-model-cm}
    \end{subfigure}
    \hfill
    \begin{subfigure}{0.49\linewidth}
        \centering
        \includegraphics[width=\linewidth]{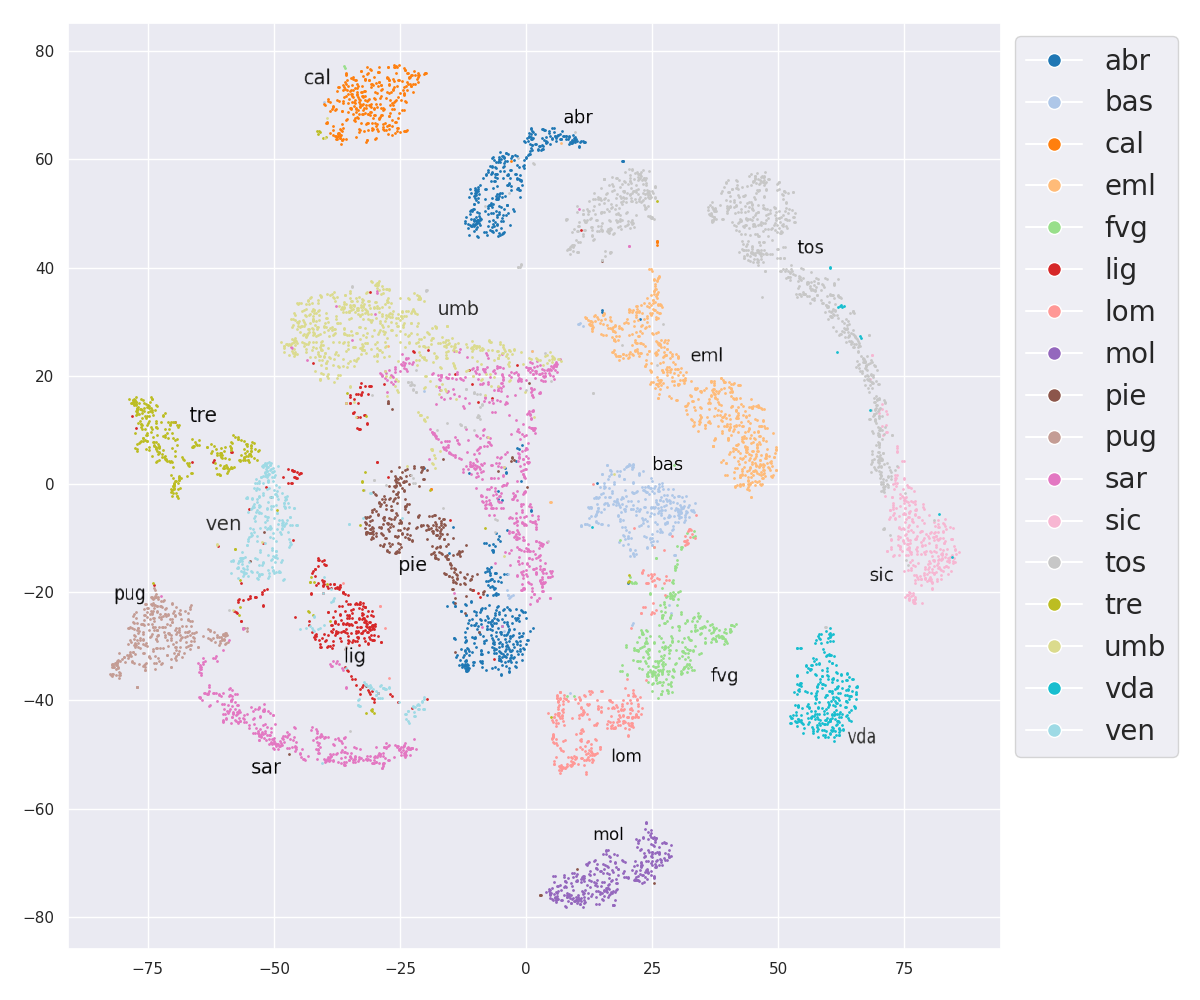}
        \caption{t-SNE visualization.}
        \label{fig:best-model-tsne}
    \end{subfigure}
    \caption{Confusion Matrix (a) and t-SNE (b) of the \texttt{XLSR-53-ITA} model w/ multi-task fine-tuning using the multi-similarity contrastive objective.}
    \label{fig:best-model}
\end{figure*}

\vspace{2mm}
\noindent \textbf{Classification task.} Figure~\ref{fig:best-model} finally displays the confusion matrix (Fig.~\ref{fig:best-model-cm}) and t-SNE projection (Fig.~\ref{fig:best-model-tsne}) of the best performing model, \texttt{XLSR-53-ITA} that underwent multi-task fine-tuning using the multi-similarity contrastive objective (i.e., second-last row of Table~\ref{table:results}).

The confusion matrix (Figure~\ref{fig:best-model-cm}) indicates that the model achieves a relatively good correlation, on average, between the true and predicted region labels.
However, there is significant confusion between specific, geographically close, region pairs, such as Basilicata (\texttt{bas}) and Calabria (\texttt{cal}), and Emilia-Romagna (\texttt{eml}) and Lombardia (\texttt{lom})\footnote{Region acronyms are provided in our repository\footref{repo_url}.}. 
This suggests proximity likely leads to more linguistic similarity that the model struggles to distinguish, especially considering there may also be linguistic variations within regions.
There is also some more unexpected confusion between regions without clear geographic explanation, like Sicilia (\texttt{sic}) with Trentino-Alto Adige (\texttt{tre}), and Liguria (\texttt{lig}) with both Trentino-Alto Adige (\texttt{tre}) and Valle d'Aosta (\texttt{vda}). 
An initial qualitative analysis of select recordings revealed that some contained a stronger presence of Standard Italian with only slight cues of the local variety. 
This suggested that some samples showed more language mixing than expected, which could make it harder to clearly distinguish between varieties.\\
\noindent
%This suggested some samples exhibited more linguistic mixing than expected, which could introduce ambiguity, complicating clear distinctions between varieties.
Overall, while the model achieves good average performance, the confusion matrix highlights where it still struggles to reliably differentiate nuances between certain challenging regions' pairs.

The t-SNE visualization (Fig.~\ref{fig:best-model-tsne}) further demonstrates the model's ability to discriminate between samples from different Italian regions. Distinct clusters are clearly formed corresponding to each region's data points. In contrast to the projections in Figures~\ref{fig:tsne-original} (the \textit{raw} \texttt{XLSR-53-ITA} model) and~\ref{fig:tsne-ms} (the version pre-trained with the MS loss), this fine-tuned model is highly effective at grouping points from the same region while separating those from other regions.
Incorporating multi-task fine-tuning with a multi-similarity contrastive objective enables this model to far surpass the original \texttt{XLSR-53-ITA} version in learning representations that properly encode relationships between samples according to their Italian regional provenance. 
This explains its superior quantitative performance on the region classification task.

Interestingly, while most clusters are tightly formed, a few regions display some overlap or dispersion of points. This suggests the model has more difficulty fully disentangling subtler differences in certain language varieties but still captures significant inter-regional variances overall. 
This aligns with previous findings in the NLP domain~\cite{diatopit}, which also observed similar phenomena, with specific regional varieties exhibiting more confounded representations than others, when analyzing textual region identification tasks.

\section{Conclusions and Future Directions}
\label{sec:conclusion}

This work presented an analysis of the linguistic variation across Italy's many regional language varieties directly from speech data. 
By leveraging the VIVALDI dataset, we assessed the performance of pre-trained speech models on automatic region identification.
Specifically, we examined how effectively different models could distinguish the regional origins of speech samples based solely on their intrinsic %linguistic
features.
The experimental results demonstrate that modern pre-trained models, particularly those fine-tuned on Italian ASR tasks, can capture meaningful differences between Italy's diverse yet closely related regional languages.
Additionally, we showed that contrastive learning objectives can enhance the discriminative ability of learned representations when applied as auxiliary training criteria during fine-tuning. 
However, even the top-performing model were confused when dealing with specific challenging region pairs. 
This indicates the intrinsic difficulty of the task. Further modeling improvements are still needed to fully disentangle the subtle linguistic variations that differentiate regional varieties.
The analysis of the confusion matrix and t-SNE projections revealed that, while many regions were distinctly clustered, some overlap remained.

Future work will address two primary directions.
First, additional data collection efforts are needed to better cover underrepresented regions and capture intra-regional diversity through multiple speakers per location.
Second, methodological advances like tailored contrastive objectives and new pooling strategies may help extract maximally informative representations from pre-trained speech models.
Addressing dataset imbalances and pushing modeling capabilities could advance the state-of-the-art on this linguistically fascinating and technologically important task.

% Future research directions will focus on collecting additional data to (i) provide better coverage of Italian regions with limited data and (ii) capture more intra-regional variations by collecting recordings from multiple speakers within the same city. This challenging yet crucial task will require collaborative efforts between researchers and local communities to collect and annotate data representing a wide range of Italian regional varieties.
% From a methodological perspective, we will investigate the design of specific fine-grained contrastive learning objectives and alternative pooling techniques to derive recording-level representations from frame embeddings. 
% Techniques such as attention pooling may be explored as they could potentially be more effective in capturing linguistic information across all frames.

% \section{CO$_2$ Emission Related to Experiments}
% Experiments were conducted using a private infrastructure, which has a carbon efficiency of 0.29 kgCO$_2$eq/kWh. A cumulative of 60 hours of computation was performed on hardware of type RTX A6000 (TDP of 300W).
% Total emissions are estimated to be 5.22 kgCO$_2$eq of which 0 percent were directly offset.
% Estimations were conducted using the \href{https://mlco2.github.io/impact#compute}{MachineLearning Impact calculator} presented in \citet{lacoste2019quantifying}.

\section{Ethical statement}
\label{sec:ethics}

This research aims to advance the understanding of linguistic diversity and promote the preservation of understudied language varieties through technology.
However, developing speech-based models also raises ethical concerns that may deserve careful consideration.

Language data reflects social and cultural norms. 
Automatic classification models could unintentionally encode harmful stereotypes or biases if data is imbalanced or fails to capture intra-group diversity, for example by not adequately representing all cities and local subgroups~\cite{koudounas2024taslp}.
Expanding data collection through local collaboration helps provide more representation of regional communities.
Also, automated speech analysis is far from perfect and may pose risks of misclassification that could impact individuals if not interpreted carefully. 
Human oversight is crucial for any research involving human data or interactions. 
We intend neither to offend any subgroups nor make claims pertaining to cultural or personal identities with this technical work.

If conducted carefully and with input from language communities, this research has the potential to aid in documenting regional varieties and raising awareness of the richness of Italy's diverse linguistic landscape.
The well-being of language communities should always be the top priority in efforts focused on language preservation.

\section{Limitations}

A key limitation of this work stems from variations in coverage across regions within the VIVALDI dataset. Some regions have relatively few or even no speech recordings included. As a result, not all areas of Italy have equal representation in the models' training and evaluation. Models may struggle more with fine-grained distinctions in under-sampled regions compared to others.
While the dataset remains extremely valuable for its scope, efforts to expand data collection from underrepresented regions or social groups would help address these limitations. \\
\noindent
Additionally, similarly to prior NLP research~\cite{ramponi2023geolingit}, this work treats regional languages as unified varieties without modeling the fine-grained linguistic variation that exists between local forms even within the same administrative region~\cite{carta_dialetti,andreose_renzi_2013}.
In some cases, linguistic variation occurs at a hyper-local level, which is not captured by coarse regional classifications. 
While useful as a first approximation, modeling linguistic variation at a finer granularity could provide greater insight by linking each city to its administrative region and geographical areas of shared linguistic roots.

\nocite{*}
\section{Bibliographical References}
\label{sec:reference}
\bibliographystyle{lrec-coling2024-natbib}
\bibliography{camera_ready}

\begin{thebibliography}{67}
\expandafter\ifx\csname natexlab\endcsname\relax\def\natexlab#1{#1}\fi

\bibitem[{Abdullah et~al.(2023)Abdullah, Shaik, and
  Klakow}]{sli_discrete_representations}
Badr~M. Abdullah, Mohammed~Maqsood Shaik, and Dietrich Klakow. 2023.
\newblock \href {https://doi.org/10.18653/v1/2023.sigtyp-1.20} {On the nature
  of discrete speech representations in multilingual self-supervised models}.
\newblock In \emph{Proceedings of the 5th Workshop on Research in Computational
  Linguistic Typology and Multilingual NLP}, pages 159--161, Dubrovnik,
  Croatia. Association for Computational Linguistics.

\bibitem[{Adams and McKenzie(2018)}]{adams2018crowdsourcing}
Benjamin Adams and Grant McKenzie. 2018.
\newblock Crowdsourcing the character of a place: Character-level convolutional
  networks for multilingual geographic text classification.
\newblock \emph{Transactions in GIS}, 22(2):394--408.

\bibitem[{Al-Tahan and Mohsenzadeh(2021)}]{al2021clar}
Haider Al-Tahan and Yalda Mohsenzadeh. 2021.
\newblock Clar: Contrastive learning of auditory representations.
\newblock In \emph{International Conference on Artificial Intelligence and
  Statistics}, pages 2530--2538. PMLR.

\bibitem[{Alumäe et~al.(2023)Alumäe, Kukk, Le, Barras, Messaoudi, and {Ben
  Kheder}}]{alumae23_interspeech}
Tanel Alumäe, Kunnar Kukk, Viet-Bac Le, Claude Barras, Abdel Messaoudi, and
  Waad {Ben Kheder}. 2023.
\newblock \href {https://doi.org/10.21437/Interspeech.2023-1790} {{Exploring
  the Impact of Pretrained Models and Web-Scraped Data for the 2022 NIST
  Language Recognition Evaluation}}.
\newblock In \emph{Proc. INTERSPEECH 2023}, pages 516--520.

\bibitem[{Andreose and Renzi(2013)}]{andreose_renzi_2013}
Alvise Andreose and Lorenzo Renzi. 2013.
\newblock \href {https://doi.org/10.1017/CHO9781139019996.009} {\emph{Geography
  and distribution of the Romance languages in Europe}}, volume~2, page
  283–334. Cambridge University Press.

\bibitem[{Ardila et~al.(2020)Ardila, Branson, Davis, Kohler, Meyer, Henretty,
  Morais, Saunders, Tyers, and Weber}]{commonvoice}
Rosana Ardila, Megan Branson, Kelly Davis, Michael Kohler, Josh Meyer, Michael
  Henretty, Reuben Morais, Lindsay Saunders, Francis Tyers, and Gregor Weber.
  2020.
\newblock Common voice: A massively-multilingual speech corpus.
\newblock In \emph{Proceedings of the Twelfth Language Resources and Evaluation
  Conference}, pages 4218--4222.

\bibitem[{Avolio(2009)}]{avolio2009lingue}
Francesco Avolio. 2009.
\newblock Lingue e dialetti d'italia.

\bibitem[{Babu and et~al.(2022)}]{xlsr128}
Arun Babu and et~al. 2022.
\newblock \href {https://doi.org/10.21437/Interspeech.2022-143} {{XLS-R:
  Self-supervised Cross-lingual Speech Representation Learning at Scale}}.
\newblock In \emph{Proc. Interspeech 2022}.

\bibitem[{Baevski et~al.(2020)Baevski, Zhou, Mohamed, and Auli}]{wav2vec2}
Alexei Baevski, Yuhao Zhou, Abdelrahman Mohamed, and Michael Auli. 2020.
\newblock wav2vec 2.0: A framework for self-supervised learning of speech
  representations.
\newblock \emph{Advances in neural information processing systems},
  33:12449--12460.

\bibitem[{Balntas et~al.(2016)Balntas, Riba, Ponsa, and
  Mikolajczyk}]{triplet_margin_loss}
Vassileios Balntas, Edgar Riba, Daniel Ponsa, and Krystian Mikolajczyk. 2016.
\newblock Learning local feature descriptors with triplets and shallow
  convolutional neural networks.
\newblock In \emph{Bmvc}, volume~1, page~3.

\bibitem[{Bartley et~al.(2023)Bartley, Jia, Puvvada, Kriman, and
  Ginsburg}]{accidental_learners}
Travis~M. Bartley, Fei Jia, Krishna~C. Puvvada, Samuel Kriman, and Boris
  Ginsburg. 2023.
\newblock \href {https://doi.org/10.1109/ICASSP49357.2023.10096407} {Accidental
  learners: Spoken language identification in multilingual self-supervised
  models}.
\newblock In \emph{ICASSP 2023 - 2023 IEEE International Conference on
  Acoustics, Speech and Signal Processing (ICASSP)}, pages 1--5.

\bibitem[{Bartoli et~al.(1995)Bartoli, Pellis, and
  Massobrio}]{istituto1995atlante}
Matteo Bartoli, Ugo Pellis, and Lorenzo Massobrio. 1995.
\newblock \emph{Atlante Linguistico Italiano}.
\newblock Istituto Poligrafico e Zecca dello Stato.

\bibitem[{Basile et~al.(2023)Basile, Musacchio, Polignano, Siciliani, Fiameni,
  and Semeraro}]{llamantino}
Pierpaolo Basile, Elio Musacchio, Marco Polignano, Lucia Siciliani, Giuseppe
  Fiameni, and Giovanni Semeraro. 2023.
\newblock Llamantino: Llama 2 models for effective text generation in italian
  language.
\newblock \emph{arXiv preprint arXiv:2312.09993}.

\bibitem[{Batter(1995)}]{vivaldi_sicilia}
Roland Batter. 1995.
\newblock Vivaldi-sicilia. documentazione sonora dei dialetti siciliani.

\bibitem[{Bellomaria et~al.(2019)Bellomaria, Castellucci, Favalli, and
  Romagnoli}]{Almawave-SLU}
Valentina Bellomaria, Giuseppe Castellucci, Andrea Favalli, and Raniero
  Romagnoli. 2019.
\newblock \href {http://arxiv.org/abs/1907.07526} {Almawave-slu: {A} new
  dataset for {SLU} in italian}.
\newblock \emph{CoRR}, abs/1907.07526.

\bibitem[{Berruto et~al.(2005)}]{berruto2005dialect}
Gaetano Berruto et~al. 2005.
\newblock Dialect/standard convergence, mixing, and models of language contact:
  the case of italy.
\newblock \emph{Dialect change. Convergence and divergence in European
  languages}, pages 81--97.

\bibitem[{Cerruti et~al.(2005)Cerruti, Regis et~al.}]{cerruti2005code}
Massimo Cerruti, Riccardo Regis, et~al. 2005.
\newblock Code switching'e teoria linguistica: la situazione italoromanza.
\newblock \emph{Italian Journal of Linguistics}, 17(1):179.

\bibitem[{Chen et~al.(2021)Chen, Chai, Wang, Du, Zhang, Weng, Su, Povey, Trmal,
  Zhang, Jin, Khudanpur, Watanabe, Zhao, Zou, Li, Yao, Wang, You, and
  Yan}]{gigaspeech}
Guoguo Chen, Shuzhou Chai, Guan-Bo Wang, Jiayu Du, Wei-Qiang Zhang, Chao Weng,
  Dan Su, Daniel Povey, Jan Trmal, Junbo Zhang, Mingjie Jin, Sanjeev Khudanpur,
  Shinji Watanabe, Shuaijiang Zhao, Wei Zou, Xiangang Li, Xuchen Yao, Yongqing
  Wang, Zhao You, and Zhiyong Yan. 2021.
\newblock \href {https://doi.org/10.21437/Interspeech.2021-1965} {{GigaSpeech:
  An Evolving, Multi-Domain ASR Corpus with 10,000 Hours of Transcribed
  Audio}}.
\newblock In \emph{Proc. Interspeech 2021}, pages 3670--3674.

\bibitem[{Chen et~al.(2020)Chen, Kornblith, Norouzi, and
  Hinton}]{chen2020simple}
Ting Chen, Simon Kornblith, Mohammad Norouzi, and Geoffrey Hinton. 2020.
\newblock A simple framework for contrastive learning of visual
  representations.
\newblock In \emph{International conference on machine learning}, pages
  1597--1607. PMLR.

\bibitem[{Conneau et~al.(2021)Conneau, Baevski, Collobert, Mohamed, and
  Auli}]{xlsr53}
Alexis Conneau, Alexei Baevski, Ronan Collobert, Abdelrahman Mohamed, and
  Michael Auli. 2021.
\newblock \href {https://doi.org/10.21437/Interspeech.2021-329} {{Unsupervised
  Cross-Lingual Representation Learning for Speech Recognition}}.
\newblock In \emph{Proc. Interspeech 2021}, pages 2426--2430.

\bibitem[{Costantini et~al.(2014)Costantini, Iaderola, Paoloni, Todisco
  et~al.}]{costantini2014emovo}
Giovanni Costantini, Iacopo Iaderola, Andrea Paoloni, Massimiliano Todisco,
  et~al. 2014.
\newblock Emovo corpus: an italian emotional speech database.
\newblock In \emph{Proceedings of the ninth international conference on
  language resources and evaluation (LREC'14)}, pages 3501--3504. European
  Language Resources Association (ELRA).

\bibitem[{Desplanques et~al.(2020)Desplanques, Thienpondt, and
  Demuynck}]{ecapa}
Brecht Desplanques, Jenthe Thienpondt, and Kris Demuynck. 2020.
\newblock \href {https://doi.org/10.21437/Interspeech.2020-2650} {{ECAPA-TDNN:
  Emphasized Channel Attention, Propagation and Aggregation in TDNN Based
  Speaker Verification}}.
\newblock In \emph{Proc. Interspeech 2020}, pages 3830--3834.

\bibitem[{Dobbriner and Jokisch(2019)}]{german_speech_dialect}
Johanna Dobbriner and Oliver Jokisch. 2019.
\newblock Towards a dialect classification in german speech samples.
\newblock In \emph{Speech and Computer: 21st International Conference, SPECOM
  2019, Istanbul, Turkey, August 20--25, 2019, Proceedings 21}, pages 64--74.
  Springer.

\bibitem[{Dunn and Wong(2022)}]{dunn-wong-2022-stability}
Jonathan Dunn and Sidney Wong. 2022.
\newblock \href {https://aclanthology.org/2022.coling-1.3} {Stability of
  syntactic dialect classification over space and time}.
\newblock In \emph{Proceedings of the 29th International Conference on
  Computational Linguistics}, pages 26--36, Gyeongju, Republic of Korea.
  International Committee on Computational Linguistics.

\bibitem[{Fu et~al.(2022)Fu, Li, Wang, Fan, Zhang, Chen, Wu, and
  He}]{fu22_interspeech}
Li~Fu, Xiaoxiao Li, Runyu Wang, Lu~Fan, Zhengchen Zhang, Meng Chen, Youzheng
  Wu, and Xiaodong He. 2022.
\newblock \href {https://doi.org/10.21437/Interspeech.2022-412} {{SCaLa:
  Supervised Contrastive Learning for End-to-End Speech Recognition}}.
\newblock In \emph{Proc. Interspeech 2022}, pages 1006--1010.

\bibitem[{Gales et~al.(2014)Gales, Knill, Ragni, and Rath}]{babel}
Mark~JF Gales, Kate~M Knill, Anton Ragni, and Shakti~P Rath. 2014.
\newblock Speech recognition and keyword spotting for low-resource languages:
  Babel project research at cued.
\newblock In \emph{Fourth International workshop on spoken language
  technologies for under-resourced languages (SLTU-2014)}, pages 16--23.
  International Speech Communication Association (ISCA).

\bibitem[{Gallipoli et~al.(2023)Gallipoli, Quatra, Cambrin, Greco, and
  Cagliero}]{gallipoli2023dante}
Giuseppe Gallipoli, Moreno~La Quatra, Daniele~Rege Cambrin, Salvatore Greco,
  and Luca Cagliero. 2023.
\newblock \href {https://ceur-ws.org/Vol-3473/paper14.pdf} {{DANTE} at
  geolingit: Dialect-aware multi-granularity pre-training for locating tweets
  within italy}.
\newblock In \emph{Proceedings of the Eighth Evaluation Campaign of Natural
  Language Processing and Speech Tools for Italian. Final Workshop {(EVALITA}
  2023), Parma, Italy, September 7th-8th, 2023}, volume 3473 of \emph{{CEUR}
  Workshop Proceedings}.

\bibitem[{Gaman et~al.(2020)Gaman, Hovy, Ionescu, Jauhiainen, Jauhiainen,
  Lind{\'e}n, Ljube{\v{s}}i{\'c}, Partanen, Purschke, Scherrer, and
  Zampieri}]{gaman-etal-2020-report}
Mihaela Gaman, Dirk Hovy, Radu~Tudor Ionescu, Heidi Jauhiainen, Tommi
  Jauhiainen, Krister Lind{\'e}n, Nikola Ljube{\v{s}}i{\'c}, Niko Partanen,
  Christoph Purschke, Yves Scherrer, and Marcos Zampieri. 2020.
\newblock \href {https://aclanthology.org/2020.vardial-1.1} {A report on the
  {V}ar{D}ial evaluation campaign 2020}.
\newblock In \emph{Proceedings of the 7th Workshop on NLP for Similar
  Languages, Varieties and Dialects}, pages 1--14, Barcelona, Spain (Online).
  International Committee on Computational Linguistics (ICCL).

\bibitem[{Giorgi et~al.(2021)Giorgi, Nitski, Wang, and
  Bader}]{giorgi2021declutr}
John Giorgi, Osvald Nitski, Bo~Wang, and Gary Bader. 2021.
\newblock Declutr: Deep contrastive learning for unsupervised textual
  representations.
\newblock In \emph{Proceedings of the 59th Annual Meeting of the Association
  for Computational Linguistics and the 11th International Joint Conference on
  Natural Language Processing (Volume 1: Long Papers)}, pages 879--895.

\bibitem[{Goebl(1989)}]{Goebl1}
Hans Goebl. 1989.
\newblock \href {https://doi.org/doi:10.1515/9783110883459-016}
  {\emph{Problèmes et méthodes de la dialectométrie}}, pages 165--184. De
  Gruyter Mouton, Berlin, Boston.

\bibitem[{Goebl(2006)}]{Goebl2}
Hans Goebl. 2006.
\newblock \href {https://doi.org/10.1093/llc/fql042} {{Recent Advances in
  Salzburg Dialectometry}}.
\newblock \emph{Literary and Linguistic Computing}, 21(4):411--435.

\bibitem[{H{\"a}m{\"a}l{\"a}inen et~al.(2021)H{\"a}m{\"a}l{\"a}inen, Alnajjar,
  Partanen, and Rueter}]{finnish_dialect_identification}
Mika H{\"a}m{\"a}l{\"a}inen, Khalid Alnajjar, Niko Partanen, and Jack Rueter.
  2021.
\newblock \href {https://doi.org/10.18653/v1/2021.emnlp-main.692} {{F}innish
  dialect identification: The effect of audio and text}.
\newblock In \emph{Proceedings of the 2021 Conference on Empirical Methods in
  Natural Language Processing}, pages 8777--8783, Online and Punta Cana,
  Dominican Republic. Association for Computational Linguistics.

\bibitem[{Kahn et~al.(2020)Kahn, Rivi{\`e}re, Zheng, Kharitonov, Xu,
  Mazar{\'e}, Karadayi, Liptchinsky, Collobert, Fuegen et~al.}]{librilight}
Jacob Kahn, Morgane Rivi{\`e}re, Weiyi Zheng, Evgeny Kharitonov, Qiantong Xu,
  Pierre-Emmanuel Mazar{\'e}, Julien Karadayi, Vitaliy Liptchinsky, Ronan
  Collobert, Christian Fuegen, et~al. 2020.
\newblock Libri-light: A benchmark for asr with limited or no supervision.
\newblock In \emph{ICASSP 2020-2020 IEEE International Conference on Acoustics,
  Speech and Signal Processing (ICASSP)}, pages 7669--7673. IEEE.

\bibitem[{Kakouros and Hiovain-Asikainen(2023)}]{northsami_dialect}
Sofoklis Kakouros and Katri Hiovain-Asikainen. 2023.
\newblock \href {https://doi.org/10.21437/Interspeech.2023-1928} {{North Sámi
  Dialect Identification with Self-supervised Speech Models}}.
\newblock In \emph{Proc. INTERSPEECH 2023}, pages 5306--5310.

\bibitem[{Kattenbusch and K{\"o}hler(2004)}]{vivaldi_sardegna}
Dieter Kattenbusch and Carola K{\"o}hler. 2004.
\newblock La sardegna nel progetto vivaldi.
\newblock \emph{GRIMALDI, LUCIA \& MENSCHING, GUIDO (a cura di), Su sardu.
  Limba de Sardigna e limba de Europa. Cagliari: Cuec}, pages 193--203.

\bibitem[{Kattenbusch et~al.(2011)Kattenbusch, Tosques, and
  Rauher}]{vivaldi_umbria}
Dieter Kattenbusch, Fabio Tosques, and Andreas Rauher. 2011.
\newblock „umbria dialettale “.
\newblock \emph{Claudia Schlaak, Lena Busse (Hg.), Sprachkontakte,
  Sprachvariation und Sprachwandel, T{\"u}bingen}, pages 443--460.

\bibitem[{Khosla et~al.(2020{\natexlab{a}})Khosla, Teterwak, Wang, Sarna, Tian,
  Isola, Maschinot, Liu, and Krishnan}]{khosla2020supervised}
Prannay Khosla, Piotr Teterwak, Chen Wang, Aaron Sarna, Yonglong Tian, Phillip
  Isola, Aaron Maschinot, Ce~Liu, and Dilip Krishnan. 2020{\natexlab{a}}.
\newblock Supervised contrastive learning.
\newblock \emph{Advances in neural information processing systems},
  33:18661--18673.

\bibitem[{Khosla et~al.(2020{\natexlab{b}})Khosla, Teterwak, Wang, Sarna, Tian,
  Isola, Maschinot, Liu, and Krishnan}]{sup_crl}
Prannay Khosla, Piotr Teterwak, Chen Wang, Aaron Sarna, Yonglong Tian, Phillip
  Isola, Aaron Maschinot, Ce~Liu, and Dilip Krishnan. 2020{\natexlab{b}}.
\newblock Supervised contrastive learning.
\newblock \emph{Advances in neural information processing systems},
  33:18661--18673.

\bibitem[{Koudounas et~al.(2023{\natexlab{a}})Koudounas, Giobergia, Benedetto,
  Monaco, Cagliero, Apiletti, and Baralis}]{koudounas2023barhotti}
Alkis Koudounas, Flavio Giobergia, Irene Benedetto, Simone Monaco, Luca
  Cagliero, Daniele Apiletti, and Elena Baralis. 2023{\natexlab{a}}.
\newblock \href {https://ceur-ws.org/Vol-3473/paper16.pdf} {ba$\rho$tti at
  geolingit: Beyond boundaries, enhancing geolocation prediction and dialect
  classification on social media in italy}.
\newblock In \emph{Proceedings of the Eighth Evaluation Campaign of Natural
  Language Processing and Speech Tools for Italian. Final Workshop {(EVALITA}
  2023), Parma, Italy, September 7th-8th, 2023}, volume 3473 of \emph{{CEUR}
  Workshop Proceedings}.

\bibitem[{Koudounas et~al.(2023{\natexlab{b}})Koudounas, {La Quatra}, Vaiani,
  Colomba, Attanasio, Pastor, Cagliero, and Baralis}]{italic}
Alkis Koudounas, Moreno {La Quatra}, Lorenzo Vaiani, Luca Colomba, Giuseppe
  Attanasio, Eliana Pastor, Luca Cagliero, and Elena Baralis.
  2023{\natexlab{b}}.
\newblock \href {https://doi.org/10.21437/Interspeech.2023-1980} {{ITALIC: An
  Italian Intent Classification Dataset}}.
\newblock In \emph{Proc. INTERSPEECH 2023}, pages 2153--2157.

\bibitem[{Koudounas et~al.(2024)Koudounas, Pastor, Attanasio, Mazzia, Giollo,
  Gueudre, Reale, Cagliero, Cumani, de~Alfaro, Baralis, and
  Amberti}]{koudounas2024taslp}
Alkis Koudounas, Eliana Pastor, Giuseppe Attanasio, Vittorio Mazzia, Manuel
  Giollo, Thomas Gueudre, Elisa Reale, Luca Cagliero, Sandro Cumani, Luca
  de~Alfaro, Elena Baralis, and Daniele Amberti. 2024.
\newblock \href {https://doi.org/10.1109/TASLP.2024.3363447} {Towards
  comprehensive subgroup performance analysis in speech models}.
\newblock \emph{IEEE/ACM Transactions on Audio, Speech, and Language
  Processing}, 32:1468--1480.

\bibitem[{{La Quatra} and Cagliero(2023)}]{BARTIT}
Moreno {La Quatra} and Luca Cagliero. 2023.
\newblock \href {https://doi.org/10.3390/fi15010015} {Bart-it: An efficient
  sequence-to-sequence model for italian text summarization}.
\newblock \emph{Future Internet}, 15(1).

\bibitem[{Lee et~al.(2023)Lee, Greenberg, Godard, Butt, Singer, Nguyen, Mason,
  and Reynolds}]{lre_2022}
Yooyoung Lee, Craig Greenberg, Eliot Godard, Asad~A Butt, Elliot Singer, Trang
  Nguyen, Lisa Mason, and Douglas Reynolds. 2023.
\newblock The 2022 nist language recognition evaluation.
\newblock \emph{arXiv preprint arXiv:2302.14624}.

\bibitem[{Liu et~al.(2022)Liu, Perera, Khong, Chng, Styles, and
  Khudanpur}]{efficient_sli_ssl}
Hexin Liu, Leibny Paola~Garcia Perera, Andy W.~H. Khong, Eng~Siong Chng,
  Suzy~J. Styles, and Sanjeev Khudanpur. 2022.
\newblock \href {https://doi.org/10.1109/JSTSP.2022.3201445} {Efficient
  self-supervised learning representations for spoken language identification}.
\newblock \emph{IEEE Journal of Selected Topics in Signal Processing},
  16(6):1296--1307.

\bibitem[{Maiden and Parry(2006)}]{maiden2006dialects}
Martin Maiden and Mair Parry. 2006.
\newblock \emph{The dialects of Italy}.
\newblock Routledge.

\bibitem[{Marini et~al.(2021)Marini, Viganò, Corbo, Zettin, Simoncini,
  Fattori, D’Anna, Donati, and Fanucci}]{IDEA}
Marco Marini, Mauro Viganò, Massimo Corbo, Marina Zettin, Gloria Simoncini,
  Bruno Fattori, Clelia D’Anna, Massimiliano Donati, and Luca Fanucci. 2021.
\newblock \href {https://doi.org/10.1109/SLT48900.2021.9383467} {Idea: An
  italian dysarthric speech database}.
\newblock In \emph{2021 IEEE Spoken Language Technology Workshop (SLT)}, pages
  1086--1093.

\bibitem[{Moisio et~al.(2023)Moisio, Porjazovski, Rouhe, Getman, Virkkunen,
  AlGhezi, Lennes, Gr{\'o}sz, Lind{\'e}n, and Kurimo}]{finnish_dataset}
Anssi Moisio, Dejan Porjazovski, Aku Rouhe, Yaroslav Getman, Anja Virkkunen,
  Ragheb AlGhezi, Mietta Lennes, Tam{\'a}s Gr{\'o}sz, Krister Lind{\'e}n, and
  Mikko Kurimo. 2023.
\newblock Lahjoita puhetta: a large-scale corpus of spoken finnish with some
  benchmarks.
\newblock \emph{Language Resources and Evaluation}, 57(3):1295--1327.

\bibitem[{Pellegrini(1977)}]{carta_dialetti}
Giovan~Battista Pellegrini. 1977.
\newblock \href {https://doi.org/https://hdl.handle.net/11168/11.318149}
  {\emph{Carta dei dialetti d'Italia}}.
\newblock Pacini, Pisa.
\newblock Illustrated, includes 1 folded geographic map (125x99 cm.).

\bibitem[{Pratap et~al.(2020)Pratap, Xu, Sriram, Synnaeve, and Collobert}]{mls}
Vineel Pratap, Qiantong Xu, Anuroop Sriram, Gabriel Synnaeve, and Ronan
  Collobert. 2020.
\newblock \href {https://doi.org/10.21437/Interspeech.2020-2826} {{MLS: A
  Large-Scale Multilingual Dataset for Speech Research}}.
\newblock In \emph{Proc. Interspeech 2020}, pages 2757--2761.

\bibitem[{Rahimi et~al.(2017)Rahimi, Baldwin, and
  Cohn}]{rahimi-etal-2017-continuous}
Afshin Rahimi, Timothy Baldwin, and Trevor Cohn. 2017.
\newblock \href {https://doi.org/10.18653/v1/D17-1016} {Continuous
  representation of location for geolocation and lexical dialectology using
  mixture density networks}.
\newblock In \emph{Proceedings of the 2017 Conference on Empirical Methods in
  Natural Language Processing}, pages 167--176, Copenhagen, Denmark.
  Association for Computational Linguistics.

\bibitem[{Ramponi(2022)}]{ramponi2022nlp}
Alan Ramponi. 2022.
\newblock Nlp for language varieties of italy: Challenges and the path forward.
\newblock \emph{arXiv preprint arXiv:2209.09757}.

\bibitem[{Ramponi and Casula(2023{\natexlab{a}})}]{diatopit}
Alan Ramponi and Camilla Casula. 2023{\natexlab{a}}.
\newblock \href {https://doi.org/10.18653/v1/2023.vardial-1.19} {{D}iatop{I}t:
  A corpus of social media posts for the study of diatopic language variation
  in {I}taly}.
\newblock In \emph{Tenth Workshop on NLP for Similar Languages, Varieties and
  Dialects (VarDial 2023)}, pages 187--199, Dubrovnik, Croatia. Association for
  Computational Linguistics.

\bibitem[{Ramponi and Casula(2023{\natexlab{b}})}]{ramponi2023geolingit}
Alan Ramponi and Camilla Casula. 2023{\natexlab{b}}.
\newblock Geolingit at evalita 2023: Overview of the geolocation of linguistic
  variation in italy task.
\newblock In \emph{Proceedings of the Eighth Evaluation Campaign of Natural
  Language Processing and Speech Tools for Italian. Final Workshop (EVALITA
  2023), CEUR. org, Parma, Italy}.

\bibitem[{Ravanelli et~al.(2021)Ravanelli, Parcollet, Plantinga, Rouhe,
  Cornell, Lugosch, Subakan, Dawalatabad, Heba, Zhong, Chou, Yeh, Fu, Liao,
  Rastorgueva, Grondin, Aris, Na, Gao, Mori, and Bengio}]{speechbrain}
Mirco Ravanelli, Titouan Parcollet, Peter Plantinga, Aku Rouhe, Samuele
  Cornell, Loren Lugosch, Cem Subakan, Nauman Dawalatabad, Abdelwahab Heba,
  Jianyuan Zhong, Ju-Chieh Chou, Sung-Lin Yeh, Szu-Wei Fu, Chien-Feng Liao,
  Elena Rastorgueva, François Grondin, William Aris, Hwidong Na, Yan Gao,
  Renato~De Mori, and Yoshua Bengio. 2021.
\newblock \href {http://arxiv.org/abs/2106.04624} {{SpeechBrain}: A
  general-purpose speech toolkit}.
\newblock ArXiv:2106.04624.

\bibitem[{Rotaru et~al.(2023)Rotaru, Ristea, and
  Ionescu}]{romanian_dialect_dataset}
Codrut Rotaru, Nicolae-Catalin Ristea, and Radu~Tudor Ionescu. 2023.
\newblock Rodia: A new dataset for romanian dialect identification from speech.
\newblock \emph{arXiv preprint arXiv:2309.03378}.

\bibitem[{Sadjadi et~al.(2018)Sadjadi, Kheyrkhah, Greenberg, Singer, Reynolds,
  Mason, and Hernandez-Cordero}]{lre_17}
Seyed~Omid Sadjadi, Timothee Kheyrkhah, Craig Greenberg, Elliot Singer, Douglas
  Reynolds, Lisa Mason, and Jaime Hernandez-Cordero. 2018.
\newblock \href {https://doi.org/10.21437/Interspeech.2018-69} {{Performance
  Analysis of the 2017 NIST Language Recognition Evaluation}}.
\newblock In \emph{Proc. Interspeech 2018}, pages 1798--1802.

\bibitem[{Santilli and Rodol{\`a}(2023)}]{camoscio}
Andrea Santilli and Emanuele Rodol{\`a}. 2023.
\newblock Camoscio: An italian instruction-tuned llama.
\newblock \emph{arXiv preprint arXiv:2307.16456}.

\bibitem[{Sanyuan and et~al.(2022)}]{wavlm}
Chen Sanyuan and et~al. 2022.
\newblock \href {https://doi.org/10.1109/JSTSP.2022.3188113} {Wavlm:
  Large-scale self-supervised pre-training for full stack speech processing}.
\newblock \emph{{IEEE} J. Sel. Top. Signal Process.}

\bibitem[{Sarni et~al.(2023)Sarni, Cumani, Siniscalchi, and
  Bottino}]{sarni23_interspeech}
Salvatore Sarni, Sandro Cumani, Sabato~Marco Siniscalchi, and Andrea Bottino.
  2023.
\newblock \href {https://doi.org/10.21437/Interspeech.2023-155} {{Description
  and analysis of the KPT system for NIST Language Recognition Evaluation
  2022}}.
\newblock In \emph{Proc. INTERSPEECH 2023}, pages 1933--1937.

\bibitem[{Sarti and Nissim(2022)}]{IT5}
Gabriele Sarti and Malvina Nissim. 2022.
\newblock It5: Large-scale text-to-text pretraining for italian language
  understanding and generation.
\newblock \emph{arXiv preprint arXiv:2203.03759}.

\bibitem[{Tosques and Castellarin(2013)}]{vivaldi}
Fabio Tosques and Michele Castellarin. 2013.
\newblock Das vivaio acustico delle lingue e dei dialetti d’italia (vivaldi).

\bibitem[{Valk and Alum{\"a}e(2021)}]{voxlingua}
J{\"o}rgen Valk and Tanel Alum{\"a}e. 2021.
\newblock Voxlingua107: a dataset for spoken language recognition.
\newblock In \emph{2021 IEEE Spoken Language Technology Workshop (SLT)}, pages
  652--658. IEEE.

\bibitem[{Van~der Maaten and Hinton(2008)}]{van2008visualizing}
Laurens Van~der Maaten and Geoffrey Hinton. 2008.
\newblock Visualizing data using t-sne.
\newblock \emph{Journal of machine learning research}, 9(11).

\bibitem[{Wang et~al.(2021)Wang, Riviere, Lee, Wu, Talnikar, Haziza,
  Williamson, Pino, and Dupoux}]{voxpopuli}
Changhan Wang, Morgane Riviere, Ann Lee, Anne Wu, Chaitanya Talnikar, Daniel
  Haziza, Mary Williamson, Juan Pino, and Emmanuel Dupoux. 2021.
\newblock \href {https://doi.org/10.18653/v1/2021.acl-long.80} {{V}ox{P}opuli:
  A large-scale multilingual speech corpus for representation learning,
  semi-supervised learning and interpretation}.
\newblock In \emph{Proceedings of the 59th Annual Meeting of the Association
  for Computational Linguistics and the 11th International Joint Conference on
  Natural Language Processing (Volume 1: Long Papers)}, pages 993--1003,
  Online. Association for Computational Linguistics.

\bibitem[{Wang et~al.(2019)Wang, Han, Huang, Dong, and Scott}]{wang2019multi}
Xun Wang, Xintong Han, Weilin Huang, Dengke Dong, and Matthew~R Scott. 2019.
\newblock Multi-similarity loss with general pair weighting for deep metric
  learning.
\newblock In \emph{Proceedings of the IEEE/CVF conference on computer vision
  and pattern recognition}, pages 5022--5030.

\bibitem[{Wieling et~al.(2014)Wieling, Nerbonne, Montemagni, and
  Baayen}]{geo_diff_tuscany}
Martijn Wieling, John Nerbonne, Simonetta Montemagni, and R.~Harald Baayen.
  2014.
\newblock \href {http://www.jstor.org/stable/24672042} {Lexical differences
  between tuscan dialects and standard italian: Accounting for geographic and
  sociodemographic variation using generalized additive mixed modeling}.
\newblock \emph{Language}, 90(3):669--692.

\bibitem[{Zampieri et~al.(2020)Zampieri, Nakov, and
  Scherrer}]{zampieri2020natural}
Marcos Zampieri, Preslav Nakov, and Yves Scherrer. 2020.
\newblock Natural language processing for similar languages, varieties, and
  dialects: A survey.
\newblock \emph{Natural Language Engineering}, 26(6):595--612.

\end{thebibliography}

\end{document}